\documentclass[journal, 10pt]{IEEEtran}
\AtBeginDocument{%
  \providecommand\BibTeX{{%
    \normalfont B\kern-0.5em{\scshape i\kern-0.25em b}\kern-0.8em\TeX}}}
\usepackage{cite}    
\usepackage{graphicx}
\usepackage{float}
\graphicspath{ {figs/} }
\usepackage{amsmath}
\usepackage{dsfont}

\usepackage[ruled,vlined]{algorithm2e}
\usepackage{xcolor}
\usepackage[font=small]{caption}
\usepackage{subcaption}
\usepackage[export]{adjustbox}
\usepackage{siunitx}
\usepackage{colortbl}
\usepackage{algorithmic}
\usepackage[thinlines]{easytable}
\usepackage{siunitx}
\usepackage[normalem]{ulem}
\usepackage{amsfonts}
\usepackage{multirow,verbatim}
\usepackage{hyperref}
\hypersetup{colorlinks,allcolors=black}

\newcommand{\prob}{\mathbf{P}}
\newcommand{\FN}{\mathtt{F}}
\newcommand{\crd}{\mathtt{C}}
\newcommand{\img}{\mathtt{I}}
\newcommand{\ldr}{\mathtt{L}}
\newcommand{\cl}{\mathtt{CL}}   %coord and LiDAR concatenation 
\newcommand{\scores}{\mathbf{s}}

\newcommand{\id}{\mathds{1}}
\SetKwComment{Comment}{$\triangleright$\ }{}

\begin{document}
\title{Deep Learning on Multimodal Sensor Data at the Wireless Edge for Vehicular Network}
\author{
    \author{
    \IEEEauthorblockN{
        Batool Salehi,
        Guillem Reus-Muns,
        Debashri Roy,
        Zifeng Wang,
        Tong Jian,\\
        Jennifer Dy,
        Stratis Ioannidis, and
        Kaushik Chowdhury\\
    }
    \IEEEauthorblockA{
        Department of Electrical and Computer Engineering\\
        Northeastern University, Boston, USA\\
        Email: \{bsalehihikouei, greusmuns, droy, zifengwang, jian, jdy, ioannidis, krc\}@ece.neu.edu
    }
}
}
\maketitle
\begin{abstract}
Beam selection for millimeter-wave links in a vehicular scenario is a challenging problem, as an exhaustive search among all candidate beam pairs cannot be assuredly completed within short contact times. We solve this problem via a novel expediting beam selection by leveraging multimodal data collected from sensors like LiDAR, camera images, and GPS. We propose individual modality and distributed fusion-based deep learning (F-DL) architectures that can execute locally as well as at a mobile edge computing center (MEC), with a study on associated tradeoffs. We also formulate and solve an optimization problem that considers practical beam-searching, MEC processing and sensor-to-MEC data delivery latency overheads for determining the output dimensions of the above F-DL architectures. Results from extensive evaluations conducted on publicly available {\color{black}synthetic and home-grown real-world} datasets reveal {\color{black}95\%} and 96\% improvement in beam selection speed over classical RF-only beam sweeping, respectively. {\color{black} F-DL also outperforms the state-of-the-art techniques by 20-22\% in predicting top-10 best beam pairs.} 
% {\color{red}TALK ABOUT THE REAL WORLD DATASET, update the numbers according to the real world data.}
\end{abstract}

% \IEEEpeerreviewmaketitle

\section{Introduction }
\label{sec:intro}
\IEEEPARstart{E}merging vehicular systems are equipped with a variety of sensors that generate vast amounts of data and require \textit{multi-Gbps} transmission rates~\cite{choi2016millimeter}. %KRC - can we put a citation here? 
These sensor inputs may be needed for safety-critical vehicle operation as well as for gaining situational awareness while in motion, which needs to be timely processed at a mobile edge computing~(MEC) center to generate driving directives.
Such a large data transfer volume at short contact times can quickly saturate the sub-6 GHz band. Thus, the millimeter-wave (mmWave) band is widely considered as the ideal candidate for vehicle-to-everything~(V2X) communications~\cite{Rasheed2020}, given the promise of 2~GHz wide channels and vast under-utilized spectrum resources in the 57-72~GHz band. However, transmission in the mmWave band has associated challenges related to severe attenuation and penetration loss. Phased arrays with directional beamforming can compensate these issues by focusing RF energy at the receiver~\cite{roh2014millimeter}. 
\begin{figure}[t]
    \centering
	\includegraphics[width=0.98\linewidth]{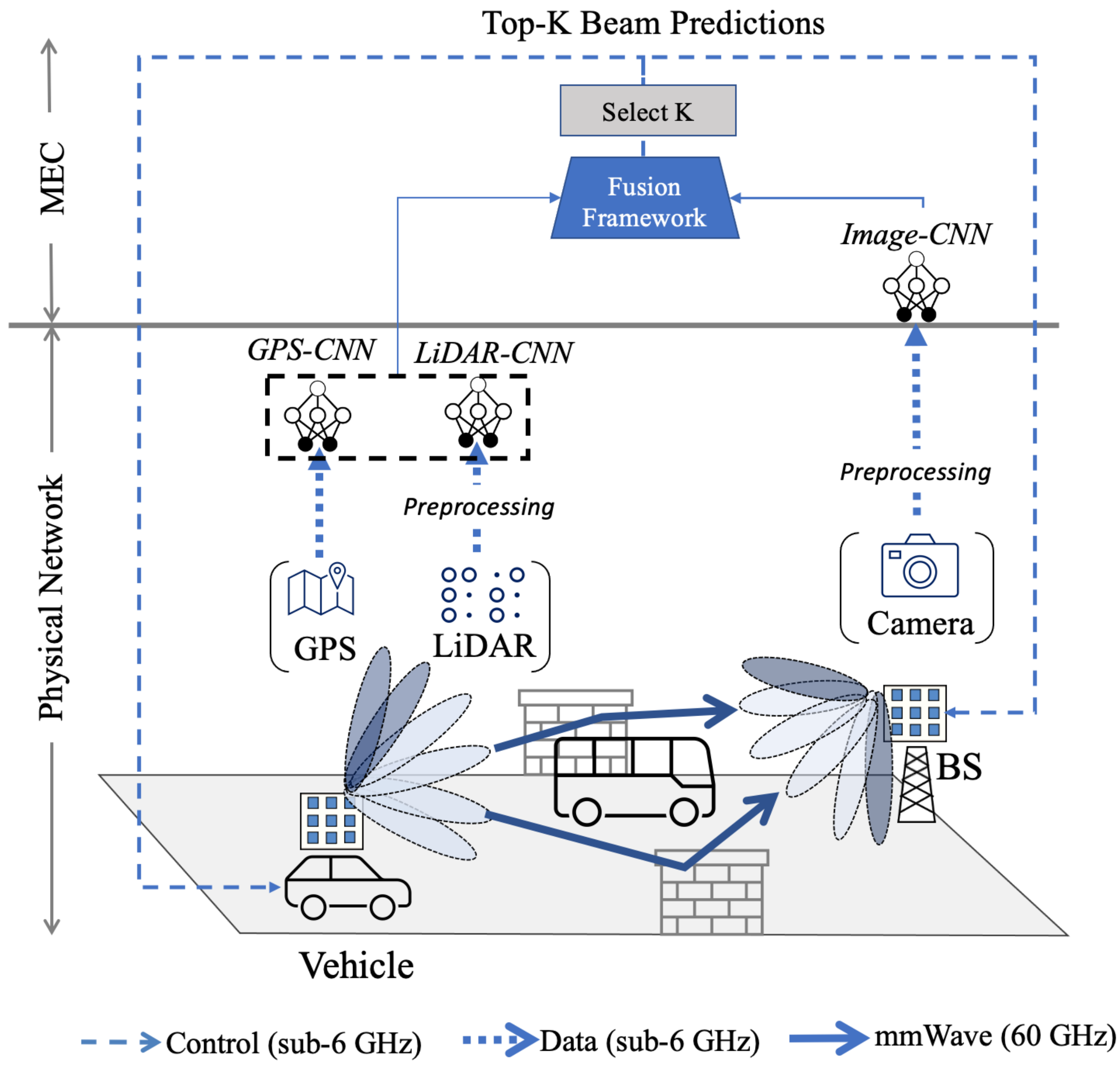}
	\caption{Our fusion pipeline exploits GPS, camera and LiDAR sensor data to restrict the beam selection to top-$K$ beam pairs.}
	\label{fig:intro_image}
	\vspace*{-10pt}
\end{figure} 
Hence, in the so called \textit{beam selection} process, the nodes on either end of the link attempt to converge to the optimal \textit{beam pairs}, where each beam pair is a tuple of transmitter and receiver beam indices, by mutually exploring the available space uniformly partitioned into discrete sectors~\cite{5262295}. However, exploring all possible beam directions in the existing IEEE 802.11ad~\cite{yaman2016reducing} and 5G New Radio (5G-NR)~\cite{giordani2018tutorial} standards can consume up to tens of milliseconds and must be repeated constantly during vehicular mobility \cite{kong2017millimeter,sanchez2020robust}. To address this problem, we propose to exploit the side out-of-band information to restrict the searching to a subset of most likely beam pair candidates. As shown in Table. \ref{fig:intro_bar_lot}, reducing the number of beam pairs from 60 to 30 significantly decreases the beam selection overhead by 50\% and 80\% for IEEE 802.11ad and 5G-NR standards, respectively.  

\subsection{Use of Sensors to Aid the Beam Selection} 
% A PARAGRAPH ON SENSORS, AND MOTIVATE THAT THOSE SENSORS ARE ABUNDANTLY AVAILABLE IN MODERN VEHICLES
Due to the directional transmissions at mmWave band, the beam selection process can be interpreted as locating the paired user or detecting the strongest reflection in the case of line of sight (LOS) and non-line of sight (NLOS) path, respectively. Hence, the location of the transmitter, receiver, and potential obstacles are the key factors in beam initialization. Interestingly, this information is also embedded in the situational state of the environment that can be acquired through monitoring sensor devices. 
\begin{table}[t]
    \centering
    \begin{tabular}{||c|c|c||} 
    \hline
    \multirow{2}{1cm}{Standard} & \multicolumn{2}{c||}{Time ($\,ms$)}\\ 
    \cline{2-3} 
    & 30 beam pairs & 60 beam pairs\\
    \hline\hline
     802.11ad & 9.09  & 18.18\\
    \hline
     5G-NR & 4.68  & 24.37\\
    \hline
    \end{tabular}
    \caption{The reduction in beam selection time while reducing the beam search space from 60 to 30 beam pairs.}
    \vspace*{-10pt}
    \label{fig:intro_bar_lot}
\end{table}
Fig.~\ref{fig:intro_image} shows our scenario of interest with a moving vehicle and a road-side base station~(BS) attempting to find the best beam pair with multiple reflectors and blocking objects. We assume the state of the environment is captured by a combination of GPS~(Global Positioning System) and LiDAR~(Light Detection and Ranging), which provides a 3-D representation of the surroundings, sensors in the moving vehicle, and a camera at the BS. We use a sub-6 GHz data channel for exchanging this sensor data between the vehicle and MEC. We then propose to use these non-RF sensor data to suggest a subset of ``top-$K$'' beam pairs and speed up the beam selection, consequently. The candidate set of selected beam pairs is communicated to both the BS and the vehicle over the sub-6 GHz control channel. After this, both the vehicle and the BS execute the standards-defined beam-searching algorithms, but only on the subset of top-$K$ suggested beam pairs.

It should be noted that, with the widespread of IoT devices, multiple sensors are now available as standard installations for the majority of electronic devices as well as fixed roadside infrastructures~\cite{gonzalez2017millimeter}.  LiDAR sensors are an indispensable part of modern vehicles that are used for either automated driving or collision avoidance~\cite{premebida2007lidar}. The GPS data are regularly collected and transmitted as part of basic safety messages frame in V2X applications~\cite{festag2015standards}, and surveillance cameras have been in use for decades with the growth of smart cities~\cite{liu2016large}. {\color{black}The Yole D\'evelopment report anticipates that the global market for GPS, radar, cameras, and LiDARs will increase from $\$67.14$ in 2020 to $\$159.6$ in 2025~\cite{yole}.}

% Note that using multiple sensor types ensures that (i) diverse aspects of the environment such as the presence of obstacles are represented in the data, and (ii) the prediction framework can flexibly adapt to situations where any number of modalities are present.
%Images provide a stable point of reference, and are able to capture the the entirety of the scenario in one-shot; (ii) LiDAR provides the crucial object detection and ranging information, and yet easily deployable over the V2I scenarios and (iii) GPS coordinates are the complement to LiDAR that enable the identification of the target vehicle. 

\vspace{-4mm}
\subsection{Deep Learning on Multimodal Sensor Data} 
%KRC- what is unclear is why is Deep ML needed? Why cannot we do this with deterministic methods? \Ok
% A PARAGRAPH WHY WE NEED DL TO OVERCOME EXPERT INTERVENE
% A PARAGRAPH CHALLENGES IN THE SYSTEM AND WIRELESS ENVIRONEMNT AND SOLUTION WITH PROPOSED FUSION

While using sensor data for out-of-band beam selection is an exciting new approach there some challenges that need to be addressed. First, since the physical environment influences signal propagation in ways that are hard to computationally model in real time, hand engineering features extracted from such sensor data that could be discriminative is infeasible, as there could be a vast multitude of reasons impacting the signal propagation. Second, a systematic approach is required to properly join the information from sensor modalities with different properties to predict the optimality of each beam pair. Note that while the beam pair can be inferred through basic geometry under ideal LOS conditions, such an approach fares poorly in scenarios with multiple reflections, such as in NLOS situations. Third, since the sensors are not all available at one site, both on the vehicle and BS, the secondary channels are required to maintain the connectivity between the vehicle and MEC. The communication constraints in these secondary channels need to be fully accounted for: the relaying cost of data exchange, especially massive LiDAR point cloud, might undermine the performance with respect to end-to-end latency. Finally, the beam search dimension $K$ is a control parameter that needs to set prior to starting the beam-searching process. Hence, an algorithm is required to select the appropriate $K$ to fully determine the system design.

Our approach directly addresses these challenges.  First, we design a fusion-based deep learning (F-DL) framework operating on all these different modalities to predict a subset of top-$K$ beam pairs that includes the globally optimal solution with high probability. Additionally, we adopt a distributed inference scheme to compress the raw data into high level extracted features at the vehicle to reduce the overhead on the wireless backchannel, accounting for end-to-end latency in the selection of the optimal beam. Finally, we take into account the prediction from our proposed F-DL framework along with mmWave channel efficiency to properly adjust the beam search space $K$, on a case-by-case basis.

% our proposed framework considers the latency in delivering data from the source sensor to the MEC - this is more relevant for the vehicle to MEC communication of LiDAR and GPS data, as the image data collected at the BS can be sent via fiber to the MEC with negligible delay. 

% Thus, our proposed multi-modal fusion technique incurs control signaling overhead over a sub-6 GHz backchannel.

% For this method to be practical, we need to rigorously analyze the conditions under which the latency introduced by the control data exchange is less than the time consumed by the deterministic approach of exhaustive search.
% We also note that the control signaling overhead is not a constant; {\color{blue}rather it is a function of the mobility rates that determine how often should sensors collect data, the volume of local data collected in each run, and user-setting for the desired $K$. -- WE ARE NOT DOING THIS.}  \removed
% KRC I think a small figure here can be helpful, to compare either the volume of data generated per sensor type (bar plot or table)... or the concept of exhaustive search vs K (restricted) beam search. This will also ensure Related Works starts on Page 3.  On second thoughts, lets think of this figure in the Sec 3 or another brief overview section that precedes it. \ok

\vspace{-4mm}
\subsection{Summary of Contributions}
Our main contributions are as follows:
\begin{itemize}
    \item We design deep learning architectures that predict the set of top-$K$ beam pairs using non-RF sensor data such as GPS, camera, and LiDAR, wherein the processing steps are split between the source sensor and the MEC. %data to guide the transmitter and receiver towards finding the best set of candidate beam pairs.
    % The architecture extracts high level representations from two sensors (GPS and LiDAR) and then fuses them in the edge devices (which are vehicles in V2I settings). These fused features are then transmitted to the MEC to be further fused with the extracted high level features of another sensor ({\em image} in this case). The outcome of overall distributed inference is finally coordinated to the edge devices or vehicles.
    {\color{black}We validate the improvement achieved by fusing available modalities versus unimodal data on a simulation as well as a home-grown real-world datatset. Our results show that fusion improves the prediction accuracy by 3.32--43.9\%. The proposed fusion network exhibits 20--22\% improvement in top-10 accuracy with respect to the state-of-the-art techniques.
    % We pledge to release the collected real-world dataset to the community upon acceptance of this paper.
    }
    
    % \item We formulate an optimization problem to appropriately select the set of $K$ candidate beam pairs, which takes into account both the standard-defined beam-searching overhead as well as the time to communicate and process data at the MEC. Thus, the control variable $K$ is not arbitrarily chosen, but tightly coupled to scenario constraints.
    \item We formulate an optimization problem to appropriately select the set of $K$ candidate beam pairs, which takes into account mmWave channel  efficiency  while  trying  to  maximizing  the alignment  probability, i.e. the case where the optimum beam pair is included within the suggested subset. Thus, the control variable $K$ is not arbitrarily chosen, but tightly coupled to scenario constraints. 
   % \item By dynamically selecting $K$, we propose a solution for achieving even lower latency while retaining the throughput, targeted to attain same beam selection accuracy. We also explore the impact of LoS and NLoS conditions in terms of beam selection accuracy.
   % \item We analyze the data/control channel bandwidth constraints and adopt a distributed inference scheme such that the backchannel occupation is minimized as much as possible, in an effort to make the approach feasible for practical deployment.
    \item We rigorously analyze the end-to-end latency of our proposed non-RF beam selection method and compare it with the state-of-the-art standard for mmWave communication, namely 5G-NR and demonstrate that the beam selection time decreases by 95--96\% on average while maintaining 97.95\% of the throughput, considering all the overhead of control/data signaling for both approaches. 
\end{itemize}

\vspace{-2mm}
\section{Related Work}
\vspace{-1mm}

%KRC- since we are targeting an ML special issue, we should separate the discussions under (i) ML-based and (ii) conventional beam selection. \OK

%We survey the most relevant works on speeding up the beam selection in mmWave bands, considering both deterministic and ML-based approaches. 

\begin{figure}
    \centering
    \includegraphics[width=0.95\linewidth]{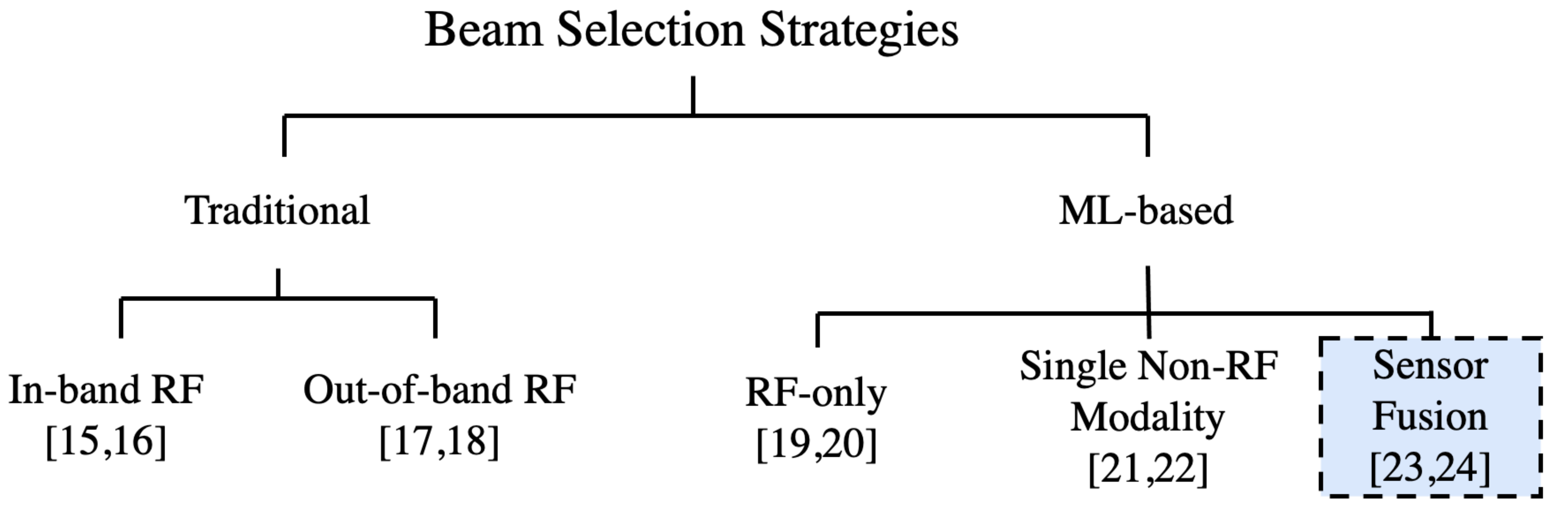}
    \caption{Deterministic and ML-aided beam selection strategies.}
    \label{fig:related_works}
    \vspace*{-10pt}
\end{figure}

Leveraging out-of-band data, both in RF and non-RF domains, can speed up the beam selection.
RF-based out-of-band beam selection is possible via simultaneous multi-band channel measurements, when there exists a mapping between mmWave and the channel state information (CSI) from the another band~\cite{sur2017wifi}. 
However, this method does not support simultaneous beamforming at both the transmitter and receiver ends. %In contrast, non-RF out-of-band beam selection can generate a mutual decision for both transmitter and receiver.
As opposed to the RF-only approach, non-RF out-of-band beam selection leverages data from different sensors and generates a mutual decision for both transmitter and receiver.
% such as  GPS, camera, and LiDAR that are available as standard installations in vehicles and fixed roadside infrastructure.
% While this is a novel approach, we need a systematic framework capable of processing all this information in a deterministic time window.  %and an analytical formulation of the information relaying overhead to the MEC, especially when the sensors are not all available at one site, e.g., both on the vehicle and base station (BS).   
Fig. \ref{fig:related_works} summarises the emphasis of this paper and different beam selection strategies.
% KRC- re-organize below \Ok
\subsection{Traditional}
\subsubsection{In-band RF}
Yang et al.~\cite{xiao2016hierarchical} adopt a hierarchical search strategy where the mmWave channel is first tested with comparatively wider beams by using a reduced number of antenna elements. The beam width is then narrows until the best beam is obtained. Wang et al.~\cite{wang2020demystifying} show that mmWave links preserve sparsity even across locations in mobile V2X scenarios. Hence, they utilize the angle of departure (AoD) to search for beams only within this range, thereby reducing beam selection overhead. 

% The proposed method devise an efficient hierarchical codebook by jointly exploiting sub-array and deactivation (turning-off) antenna processing techniques and closed-form expressions are provided to generate the codebook. 
\subsubsection{Out-of-band RF}
%This strategy employs channel state information at lower/higher frequency bands to aid the beam selection in mmWave band. Steering with eyes closed 
Steering with eyes closed~\cite{nitsche2015steering} exploits  omni-directional transmissions within the legacy 2.4/5 GHz band to infer the LOS direction between the communicating devices to speed up the mmWave beam selection. González-Prelcic et al.~\cite{gonzalez2016radar} exploit the side information derived from RAdio Detection And Ranging (RADAR) data to adapt the beams in a vehicle to infrastructure  network, where a compressive covariance estimation approach is used to establish a mapping between RADAR and mmWave bands. 

\subsection{ML-based}
\subsubsection{RF-only}
He et al.~\cite{he2018deep} design a deep learning based channel estimation approach using iterative signal recovery, wherein the channel matrix is regarded as a noisy 2D natural image. Learnt denoising-based approximate message passing (LDAMP) neural networks are applied on the input for channel estimation. Hashemi et al.~\cite{hashemi2018efficient} model the mmWave beam selection as a MAB (Multi-armed bandit) and use the reinforcement learning to maximize the directivity gain (i.e., received energy) of the beam alignment policy. % within a time period. 

\subsubsection{ML using single non-RF modality}
Va et al.~\cite{va2017inverse} consider a setting where the location of all vehicles on the road, including the target receiver, is used as input to a machine learning algorithm to infer the best beam configuration. Vision-aided mmWave beam tracking in \cite{alrabeiah2020viwi} models a dynamic outdoor mmWave communication setting where the sequence of previous beams and visual images are used to predict future best beam pairs.

% While the majority of current literature use synthetic dataset, \cite{MASS2020} deploy a testbed using National Instruments radio at 60 GHz and camera images to predict the best beam configuration.
\subsubsection{ML with sensor fusion}
% Similarly, fusion based solutions for beam-sweeping from multiple data modalities have also been explored in literature.
The proposed setting by Klautau et al.~\cite{klautau2019} and Dias et al.~\cite{dias2019position} comes closest to ours with GPS and LiDAR being used as the side information for LOS detection and also reducing the overhead in a vehicular setting. 

{\color{black} The state-of-the-art~\cite{klautau2019,dias2019position} does not consider the deep learning based fusion for more than two non-RF modalities to fully exploit the latent features within the data. The GPS coordinates are only used in the preprocessing pipeline to identify the target receiver. There also has not been any effort to decouple the expert knowledge for dynamically reducing the beam search space depending on specific user constraints. Our proposed method exploits a customized deep learning fusion approach that is carefully designed to maximize the beam selection accuracy. Moreover, completed by an algorithm that automatically chooses a dynamic subset of beam pairs, our method can run end-to-end without any hand engineering.}

\section{System Model and Overview}
\label{sec:systemModel}
In this section, we first review classical beam selection and discuss it's limitations. We then propose to use non-RF data from multiple sensors to facilitate-- and accelerate--beam selection. 

% Table~\ref{tab:notation} summarizes our notation. %We then exploit the properties of different sensors to make optimized deep learning architectures for each sensor type. %Finally, we describe the operation of the overall, end-to-end system.

\subsection{Beam Selection Problem Formulation}
\label{sec:problem_formulation_802.11ad}
We denote the codebook of transmitter and receiver radios:
\begin{equation}
    C_{Tx}=\{t_1,\dots,t_M\},~~  C_{Rx}=\{r_1,\dots,r_N\},
\end{equation}
where $M,N$ are the number of transmitter and receiver codebook elements, respectively. Each element of the codebook represents a particular beam orientation that can be utilized by the radio.
% that corresponds to a set of beam weights associated to each element of antenna array. 
Thus, the set of all possible beam pairs $\mathcal{B}$ is:
\begin{equation}
    \mathcal{B} = \{(t_{m},r_{n})|t_{m}\in C_{Tx},r_{n}\in C_{Rx}\},
\end{equation}
with $|\mathcal{B}|=M\times N $. For a specific beam pair $(t_m,r_n)$, the normalized signal power is obtained as:
\begin{equation}
    y_{(t_m,r_n)} = |w_{t_m}^H ~\mathbb{H}~w_{r_n}|^2,
\end{equation}
where $\mathbb{H}\in \mathbb{R}^{M\times N}$ is the channel matrix  and $H$ is the conjugate transpose operator. The weights $w_{t_m}$ and $w_{r_n}$ indicate the corresponding beam weight vectors associated with the codebook element $t_m$ and $r_n$, respectively ($|w_{t_m}|=M,|w_{r_n}|=N$). The goal of the beam selection process is to identify the best beam configuration, $(t^*,r^*)$, that maximizes the normalized signal power, given by:
\begin{equation}
    (t^*,r^*) = \underset{1\leq m \leq M , 1\leq n \leq N }{\arg\max}~y_{(t_{m},r_{n})}.
    \label{eq:argmax_q}
\end{equation}
In classical beam selection, such as the approach defined in the IEEE 802.11ad~\cite{nitsche2014ieee} and 5G-NR \cite{giordani2019standalone} standards, the transmitter and receiver sweep all beam pairs $(t_m,r_n) \in \mathcal{B}$ sequentially in order to select the best beam pair.

\subsection{Subset Selection}
\label{Sec:sub-set-selection}
While exhaustive searching through all candidate options ensures the beam alignment, the typical time to complete the entire procedure is in the order of $\sim$10 ms for IEEE 802.11ad~\cite{yaman2016reducing} and $\sim$5 ms for 5G-NR~\cite{giordani2018tutorial} with only 30 beam pairs, respectively. To address this, we propose an \emph{out-of-band beam selection framework} that uses out-of-band data to identify a subset of candidate beams, which are subsequently swept to select the one that maximizes the normalized signal power. More specifically, the key algorithmic component of our system amounts to proposing a means for identifying a subset $\mathcal{B}_K\subseteq \mathcal{B}$ of $K$ beam pairs such that $(t^*,r^*) \in \mathcal{B}_K$ with high probability. Formally, assuming that we have a probability distribution for the optimal pair $(t^*,r^*)$, we wish to find:
\begin{equation}
    \mathcal{B}_K = 
     \underset{A\subseteq \mathcal{B}, |A|=K}{\arg\max}~\prob{((t^*,r^*)\in A}).
    \label{eq:fast_p}
\end{equation}
Having obtained $\mathcal{B}_K$, we then restrict the search for the optimal pair to this set. Our solution uses  a neural network to leverage out-of-band data to determine the probability distribution $\prob$. Parameter $K$ establishes a trade-off between throughput performance, obtained by the best beam in $\mathcal{B}_K$, and latency, as a larger $K$ results in more processing time to search through the candidate options. Thus, our end-to-end design includes a means for appropriately determining $K${\color{black}, where the boundary condition of $K= 1$ represents selecting the optimal beam pair.} 

{\color{black} Overall, this auxiliary parameter $K$ enables the users to adjust the system according to their specific constraints on establishing a low-latency or ultra-reliable communication. Moreover, it gives the flexibility to analyze the adjacent beam patterns with relatively closer performance or irregular radiation patterns under NLOS conditions.}
%Thus, the chosen value for $K$ connects the F-DL architecture design to the standard-defined beam selection overhead. 

% \subsection{System Description}
% \label{sec:system_des}
% Our proposed scheme runs in four setps as follows. First, the sensors at the vehicle collect GPS and LiDAR data, whereas the camera at the BS captures visual information of the environment. The collected raw data is then preprocessed. Second, the vehicle shares the extracted features of coordinates and LiDAR sensor modalities with the MEC over the sub-6~GHz data channel. Notice that instead of sending the raw sensor data, only the high level extracted features generated by each single-modal network are shared. This approach avoids sharing unnecessary amounts of data and helps mitigating potential privacy concerns. Third, given the extracted features of all three modalities at MEC, we use our proposed fusion network to suggest a set of top-$K$ candidates for sweeping. The subset of $K$ beam pair is shared to the vehicle over the sub-6~GHz control channel. Finally, the beam sweeping runs at mmWave band (60~GHz) in a reduced search space of selected top-$K$ candidates to select the best beam pair and establish the link.

\subsection{System Overview}
\label{sec:problem_formulation_sensor}
Overall our framework consists of three main components. %as follows.
\begin{itemize}
    \item \textbf{Data Preprocessing:}  
    For the collected data to be effective, it is crucial to mark the transmitter, target receiver, and blocking objects. Thus, we exploit the prepossessing step described in Sec.~\ref{sec:preprocessing} for image and LiDAR.
    \item \textbf{Beam Prediction using Fusion-based Deep Learning:} Given the multimodal sensor data, we design a F-DL architecture that predicts the optimality of each beam pair. Our approach consists of custom-designed feature extractors for each sensor modality, followed by a fusion network that joins the information for the final prediction. Our proposed fusion approach is presented in Sec.~\ref{sec:proposed_fusion}.
    \item \textbf{Top-$K$ Beam Pair Construction:} We select, the beam search space dimension, $K$ by defining an optimization problem (see Sec. \ref{sec:dynamic_k_selection}) that takes into account the mmWave channel efficiency and probability of including the globally optimum beam pair. 
\end{itemize}

In summary, our proposed beam selection approach runs in four steps end-to-end. First, the sensors at the vehicle collect GPS and LiDAR data, and the camera at the BS captures an image. The collected raw data is then {\em preprocessed on site}. Second, having the feature extractors of GPS and LiDAR being deployed at the vehicle, the high level features are generated and {\em shared with the MEC over the sub-6~GHz data channel}. This approach avoids sharing unnecessary amounts of data and helps mitigating potential privacy concerns. The high-level features of the image are generated in parallel. Third, given the extracted features of all three modalities at MEC, our method suggest a set of top-$K$ candidates for sweeping. The subset of $K$ beam pair is {\em shared with the vehicle over the sub-6~GHz control channel}. Finally, the {\em beam sweeping runs at mmWave band (60~GHz)} in a reduced search space of selected top-$K$ candidates to select the best beam pair and establish the link.

\subsection{Sensor Modalities}
The details of the three sensor modalities are given below:
\begin{itemize}
    \item \textbf{GPS:} This sensor generates readings in the decimal degrees (DD) format, where the separation between each line of latitude or longitude (representing ${1}^{\circ}$ difference) is expressed as a float number with 5 digit precision. Each measurement results in two numbers that together pinpoints the location on the earth’s surface. We do not assume any satellite link outages due to terrain or man-made structures.
    \item \textbf{Image:} This sensor captures still RBG images of the environment. Although images allow comprehensive environmental assessment, they are impacted by low-light conditions and obstructions (such as a different vehicle in the LOS path)
    \item \textbf{LiDAR:} This sensor generates a 3-D representation of the environment by emitting pulsed laser beams. The distance of each individual object from the origin (i.e., the sensor location) is calculated based on reflection times. The raw LiDAR point clouds are data intensive~($\sim$1.5 Mb for sparse settings), necessitating processing at the vehicle itself.
\end{itemize}
\section{Data Preprocessing}
\label{sec:preprocessing}
% Note that it not feasible to distinguish between the target receiver and other vehicles by only relying on raw camera and LiDAR data, which undermines the prediction performance. To address this issue, in this section, 
In this section, we describe our preprocessing pipeline for image and LiDAR.

% Since each GPS sample is a tuple of floating point numbers, we do not need to compress this further. 
% In this section, we describe our preprocessing pipeline for image and LiDAR that extracts relevant features.  Since each GPS sample is a tuple of floating point numbers, we do not need to compress this further. 

% a. By preprocessing, we arrange the input in a refined presentation where the relevant information to our task (e.g., location of obstacles) is highlighted.
\subsection{Processing Images}
\label{sec:preprocessing_img}
% The raw images collected at the BS include the target receiver along with other vehicles. We note that using the entire image in its original state fails to distinguish the importance of certain objects in the image. Although the presence of walls and buildings impact the best beam selection, they form a static background that can be removed. Additionally, our learning architecture distinguishes between vehicle types, which also helps to separate the spatial boundaries of multiple vehicles and other objects in the same frame. 
% In summary, our approach (i) removes the static background,
% (ii) detects multiple vehicle types present in the same scene, (iii) separates the receiver and obstacle regions. To preprocess raw images, we employ a multi-object detection approach that enables us to flexibly handle the input images. Since the focus of this paper is not directly on image processing, we include details of our  custom-designed approach in Appendix \ref{appendix:imgae_preprocessing}. The output of this image preprocessing step is the \textit{bit map} of the raw input camera image and it serves as the input to our fusion pipeline.
The raw images collected at the BS provide a snapshot of the present objects in the scene. In this case, it is crucial to detect the region of the target receiver among other vehicles that correspond to the blocking objects. Hence, we design a preprocessing step as follows. First, we employ a multi-object detection approach that enables us to flexibly distinguish the spatial boundaries of different vehicle types in the same frame. Second, given the type of target vehicle, we separate the region of the target receiver and blocking vehicles. On the other hand, the background with static walls and buildings is invariant over different scenes and consequently does not affect the decision and can be further removed. In summary, our approach
(i) detects multiple vehicle types present in the same scene, (ii) separates the receiver and obstacle regions, and (i) removes the static background. Since the focus of this paper is not directly on image processing, we include details of our custom designed approach in Appendix \ref{appendix:imgae_preprocessing}. The output of this image preprocessing step is the \textit{bit map} of the raw input camera image, and it serves as the input to our fusion pipeline.
%#################################################
%###################LIDAR########################
%#################################################
\subsection{Processing LiDAR Point Clouds}
\label{sec:preprocessing_lidar}
\begin{figure}[t!]
    \centering
    \includegraphics[width=\linewidth]{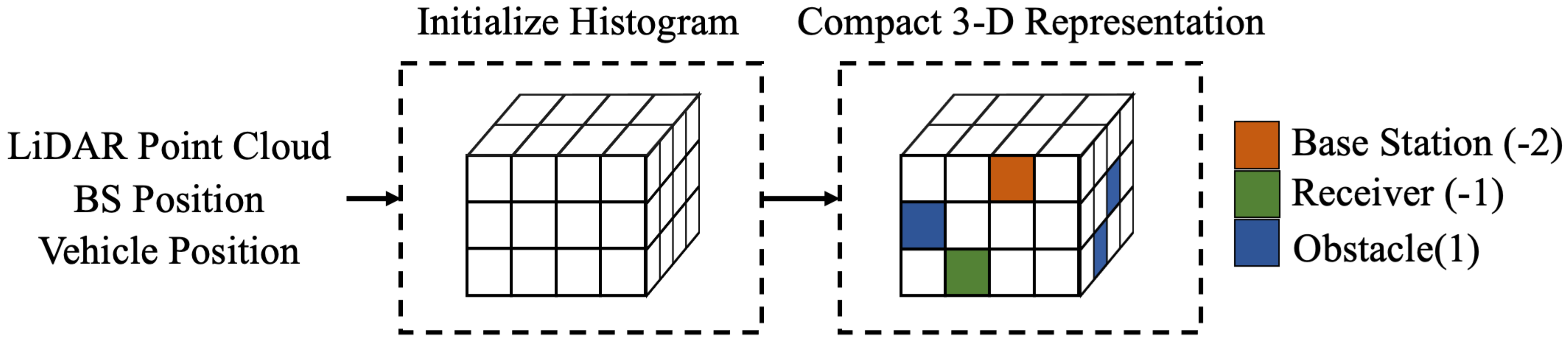}
    \caption{The LiDAR preprocessing pipeline.}
    \label{fig:LIDAR_feature_ext}
    \vspace*{-10pt}
\end{figure}
The raw LiDAR point cloud is a collection of $(x,y,z)$ points that correspond to the location of detected objects in the environment. Directly exploiting the raw point cloud (with varying number of points depending on traffic density) not only comes with huge computational cost but also raises ML architecture design challenges as the input to a neural network must be preferably fixed in size. Hence, we use a preprocessing step as shown in Fig. \ref{fig:LIDAR_feature_ext} first proposed in~\cite{klautau2019} that considers a limited spatial zone for each axis. This space corresponds to coverage range of BS and is  denoted as $(X_{\min},X_{\max})$, $(Y_{\min},Y_{\max})$, and $(Z_{\min},Z_{\max})$. Then, we construct a 3-D histogram that corresponds to a quantized 3-D representation of the space. The histogram bin size along the three spatial dimensions $(b_x,b_y,b_z)$ can be set based on desired resolution. The LiDAR point clouds lie in the corresponding bins of the histogram based on their location. Since the BS is fixed in our setting, it always occupies the same cell of the histogram with indicator ($-2$). The corresponding cell of the target receiver is also acquired with GPS data and indicated with ($-1$). The remaining elements are mapped to the corresponding histogram elements with ($1$), which implies the presence of obstacles. This leads to a 
compact 3-D representation of the environment that we use as input for our pipeline. 
% \section{Fusion based Beam Prediction}
\section{Beam Prediction using fusion-based \\ deep learning}
\label{sec:proposed_fusion}
\begin{figure}
    \centering
	\includegraphics[width=\linewidth]{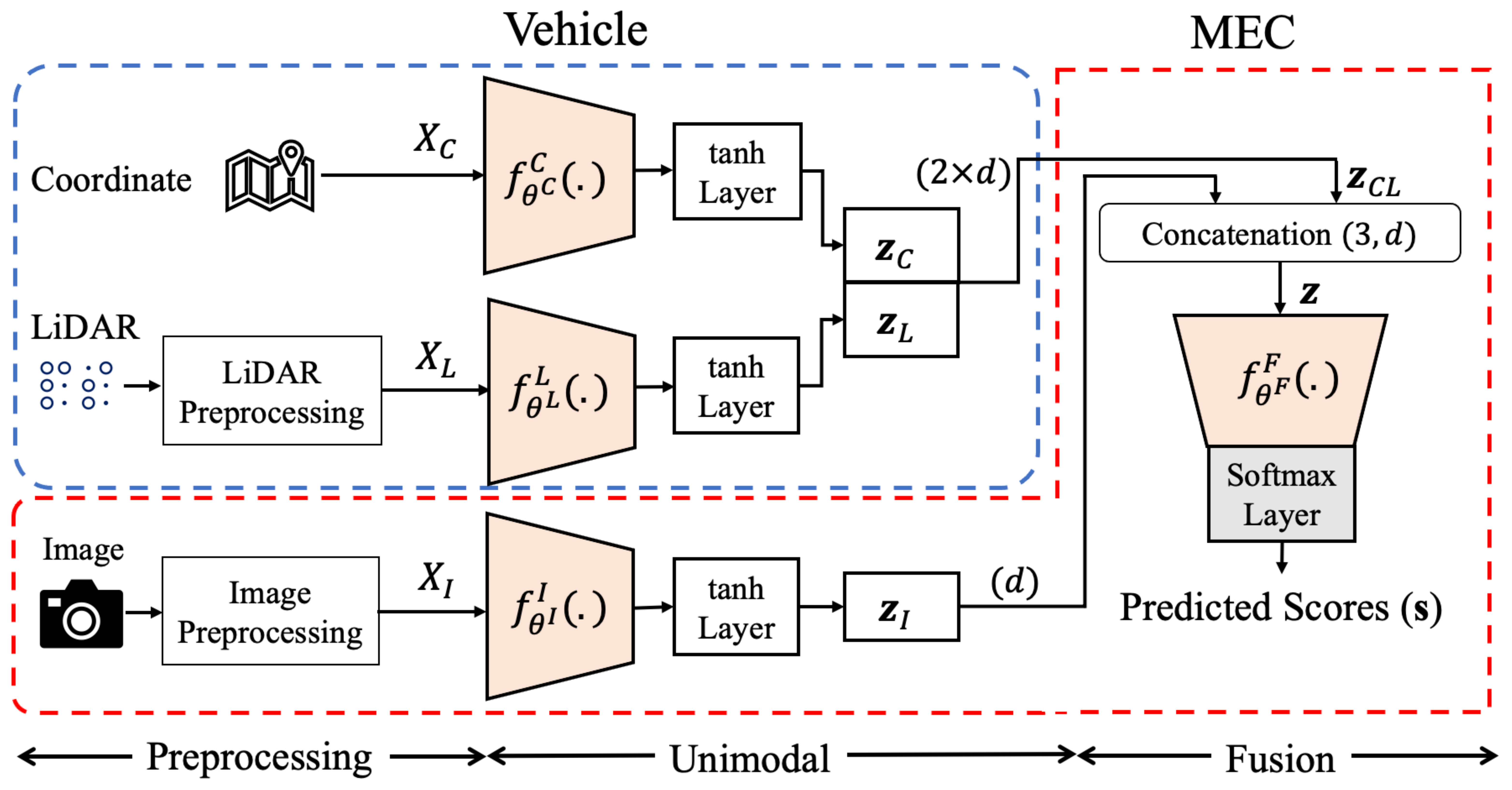}
	\caption{Proposed fusion framework. In the training phase, the pipeline is trained offline, and during the distributed inference, the trained model is disseminated over the system.}
	\vspace*{-10pt}
	\label{fig:fusion_pipeline}
\end{figure}

In the second step of our proposed framework, we design a multimodal data fusion pipeline to combine the available sensing modalities together and predict the optimality of each beam pair. First, we describe the methodology for training the fusion pipeline, followed by the proposed distributed inference approach as shown in Fig. \ref{fig:fusion_pipeline}.
\subsection{Training Phase}
We define the data matrices for GPS, LiDAR and images as: $X_{\crd}\in \mathbb{R}^{N_t\times 2}, X_{\ldr}\in \mathbb{R}^{N_t\times d_0^\ldr \times d_1^\ldr \times d_2^\ldr}, X_{\img}\in \mathbb{R}^{N_t\times d_0^\img \times d_1^\img}$, respectively, where $N_t$ is the number of training samples. Furthermore, $(d_0^\ldr \times d_1^\ldr \times d_2^\ldr)$ and $(d_0^\img \times d_1^\img)$ give the dimensionality of preprocessed LiDAR and image data, while the GPS  coordinate has 2 elements. {\color{black}We consider the label matrix $Y\in \{0,1\}^{N_t\times\mathcal{|B|}}$ to represent the one-hot encoding of $\mathcal{B}$ beam pairs,  where the optimum beam pair is set to $1$, and rest are $0$ as per Eq.~\eqref{eq:argmax_q}.} {\color{black} As mentioned in Sec.~\ref{sec:problem_formulation_802.11ad}, we have one optimal beam pair per sample, so we opted for one-hot encoding which enables having just one class per sample.
% , unlike the multi-hot encoding where multiple classes may be present in one sample.
} 
Overall, we design a fusion framework to combine different data modalities that contains two main components: (i) base unimodal networks and (ii) the fusion network. 

\noindent \\$\bullet$ {\bf Base Unimodal Neural Network:}
\label{sec:single_network} We use the base unimodal neural network to (i) benchmark the performance of our fusion-based approach with respect to what can be achieved using only a single sensor type, and (ii) extract latent features from the penultimate~(second last) layer of each that we use as input to our fusion network. 

A deep neural network (DNN) can be considered as a combination of a non-linear feature extractor followed by a softmax classifier, i.e.,  the first layer until the penultimate layer of the DNN constitute the feature extractor \cite{wang2020open}. The feature extractor maps an input to a point in a multi-dimensional space called as the latent embedding space. The dimension of this high-level data representation is equal to the number of neurons in the penultimate layer. Then, in the final layer, the  softmax activation function maps the high level representation of input data to a probability distribution over classes. As a result, the penultimate layer captures the unique properties of input data through a latent embedding space that is the key to making the final decision. 

In this work, we propose to use the output of unimodal feature extractors as the high level data representation of each sensor modality. We assume that the penultimate layer of all three unimodal networks has $d$ neurons. As a result, each sensor modality sample input maps to a vector with dimension $d$ after passing through the feature extractors. We denote the feature extractor of each modality as $f_{\theta^\crd}^\crd$, $f_{\theta^\ldr}^\ldr$ and $f_{\theta^\img}^\img$ for coordinate, LiDAR, and image data, respectively, each parametrized by weight vectors $\theta^m$, for $m\in\{\crd,\ldr,\img\}$. We refer to the output of these feature extractors as the latent embedding of each modality. Formally,%
\begin{subequations}%
    \begin{align}
     & \mathbf{z}_{\crd} = f_{\theta^\crd}^\crd(X_{\crd}),~~~~~~f_{\theta^\crd}^\crd:\mathbb{R}^{2} \mapsto \mathbb{R}^d \\
     & \mathbf{z}_{\ldr} = f_{\theta^\ldr}^\ldr(X_{\ldr}),~~~~~~f_{\theta^\ldr}^\ldr:\mathbb{R}^{d_0^\ldr \times d_1^\ldr \times d_2^\ldr} \mapsto \mathbb{R}^d\\
     & \mathbf{z}_{\img} = f_{\theta^\img}^\img(X_{\img}),~~~~~~f_{\theta^\img}^\img:\mathbb{R}^{d_0^\img \times d_1^\img} \mapsto \mathbb{R}^d
    \end{align}
\label{eq:feature_extrs}%
\end{subequations}
where $\mathbf{z}_{\crd}$, $\mathbf{z}_{\ldr}$ and $\mathbf{z}_{\img}$ show the extracted latent embeddings for input data $X_{\crd}$, $X_{\ldr}$ and $X_{\img}$, respectively. We then apply a \textit{tanh} activation on extracted latent features to regularize them in a range [-1, 1]. Note that the input to the base unimodal networks may contain negative values, which motivates the choice of \textit{tanh} as the regularization function.

\noindent\\ $\bullet$  {\bf Fusion Neural Network:}
\label{sec:fusion_network} Each of the modalities capture different aspects of the environment. For instance, the GPS coordinates provide the precise location of the target receiver but {\color{black}it is blind to the shifts in the other objects in the environment and} fails to provide any information about the dimensions of the vehicles. %One the other hand, the input data is prone to errors that might misguide the prediction. For example, the 
LiDAR accuracy degrades in bright sunshine with many reflections \cite{heinzler2019weather}. %KRC- is this well know, else put a citation?
Hence, fusing different modalities can compensate for the {\color{black}partial} or inaccurate information and increase the robustness of the prediction.

% Furthermore, feature-level fusion helps in the case of errors in the input data. For example, LiDAR accuracy degrades in high sun reflections, which might generate unassertive decision or close competing classes at output layer of the learning model. Eventually, availability of multiple modalities to perform the same task increases the robustness of the prediction.
% robust error
% compensate for missing
% not so sure

% Given the latent feature embedding of all modalities, we propose a fusion approach as follows. First, we apply a \textit{tanh} activation function on extracted latent features to regularize them in a range [-1, 1].
Given the latent feature embedding of all modalities, we propose a fusion approach as follows: We explore that feature concatenation is an effective strategy for feature-level fusion in machine learning~\cite{MFAS_CVPR}. Hence, our proposed fusion method is comprised of concatenation of latent feature embedding from each unimodal network to account for all sensor modalities, simultaneously. Thus, given $\mathbf{z}_{\crd}$, $\mathbf{z}_{\ldr}$ and $\mathbf{z}_{\img} \in \mathbb{R}^d$, we first concatenate them and generate the combined latent feature matrix $\mathbf{z}$ as:
\begin{equation}
    \mathbf{z} = [\mathbf{z}_{\crd} ; \mathbf{z}_{\ldr}: \mathbf{z}_{\img}]\in \mathbb{R}^{3\times d}. 
    \label{eq:combined_latent_em}
\end{equation}
Moreover, using multiple layers after concatenation of extracted features allows our fusion architecture to learn about the relevance of modalities altogether, and therefore, it intelligently assigns higher weights to the features of the more relevant modalities. We pass the combined latent feature matrix $\mathbf{z}$ to another convolutional neural network~(CNN) that we refer as {\em fusion network} to properly learn the relation of extracted latent embedding and the corresponding optimum beam pair. We denote the fusion network as $f_{\theta^\FN}^\FN(.)$. Finally, we use a softmax activation function to predict the optimality of each beam pair as:
\begin{equation}
\label{eq:fusion_network}%
 \scores = \sigma (f_{\theta^\FN}^\FN(\mathbf{z})),~~~~~f_{\theta^\FN}^\FN:\mathbb{R}^{3\times d} \mapsto \mathbb{R}^{|\mathcal{B}|} 
\end{equation}
where $\sigma$ denotes the softmax activation function defined as $\sigma(\mathbf{x})_i=\frac{e^{x_i}}{\sum_{j=1}^{|\mathcal{B}|} e^{x_j}}, i\in\mathcal{B}$, and $\scores = [s_i]_{i\in\mathcal{B}} \in \mathbb{R}^{|\mathcal{B}|}$
% \begin{equation}
%     \scores = [s_i]_{i\in\mathcal{B}} \in \mathbb{R}^{|\mathcal{B}|}
% \end{equation} 
indicates the predicted score of each beam pair. Note that $\scores$ forms a probability distribution, with $s_i=\prob((t^*,r^*)=i)$, $i\in\mathcal{B}$. We train this network  offline using a cross-entropy penalty, over data in which the optimal $(t^*,r^*)$ pair is one-hot encoded. %Given the multi-modal dataset including $X_{coord}$, $X_{img}$, $X_{lidar}$ and one-hot encoding of beam pairs $Y$, we conduct an offline training approach on the fusion framework shown in Fig. \ref{fig:fusion_pipeline}.

% In summary, the proposed fusion framework maps the input data of single modalities to a high level representation, that is generated by the base single-modal feature extractors as defined in Eq. \eqref{eq:feature_extrs}. The extracted latent embeddings are then concatenated and arranged as a combined latent embedding matrix that is then passed to the fusion network to infer the probability of each beam pair, Eq. \eqref{eq:fusion_network}.

\subsection{Distributed Inference Phase}
Unlike the training phase that occurs offline, the inference needs to occur real-time. To that end, the MEC receives  instantaneous data from three sensor modalities, which is passed to the trained fusion pipeline for predicting the top-$K$ beam pairs. Since the sensors are not co-located, to accelerate inference, we distribute the ML architecture taking account the limitations of the control channel delivering the sensor data to the MEC. Our distributed inference scheme is illustrated in Fig.~\ref{fig:intro_image}. The trained base unimodal networks for GPS coordinates and LiDAR are deployed at the vehicle to locally generate the high level latent embeddings $\mathbf{z}_{\crd}$ and $\mathbf{z}_{\ldr}$. The extracted features are then concatenated as $ \mathbf{z}_{\cl} = [\mathbf{z}_{\crd} ; \mathbf{z}_{\ldr}]\in \mathbb{R}^{2\times d}$ and sent over the sub-6 GHz data channel. %KRC- I thought we said, we don't do anything to GPS?
%\begin{equation}
 %   L_{cl} = [L_{coord} ; L_{lidar}]\in \mathbb{R}^{2\times d}. 
%\end{equation}
Similarly, the base (unimodal) network of the image generates the features for this modality at the BS, which is then combined with $\mathbf{z}_{\cl}$ at the MEC as $\mathbf{z} = [\mathbf{z}_{\cl}; \mathbf{z}_{\img}]\in \mathbb{R}^{3\times d}$.
%\begin{equation}
%    L = [L_{cl}; L_{img}]\in \mathbb{R}^{3\times d}. 
%\end{equation}
Note that this methodology results in the same combined latent feature matrix $\mathbf{z}$ as Eq. \eqref{eq:combined_latent_em}, we  analyze the improvement in end-to-end latency with this distributed inference approach in Sec.~\ref{sec:dist_inf}. Finally, given the latent feature embedding of all modalities available at the MEC, we use the fusion network, $f_{\theta^\FN}^\FN (\cdot)$, followed by a softmax activation to predict the score of each beam pair according to Eq.~\eqref{eq:fusion_network}. Fig.~\ref{fig:fusion_pipeline} depicts the dissemination of the fusion pipeline over the system.

% ########################################
% ################NEW SECTION################
% ########################################

\section{Top-$K$ Beam Pair Construction}
\label{sec:dynamic_k_selection}

The proposed fusion pipeline outputs a softmax score for each of the possible beam pairs given the different sensor modalities. Recall that our goal is to identify a subset of beam pairs $\mathcal{B}_K$ such that $(t^*,r^*) \in \mathcal{B}_K$ with high probability. We describe in this section how the neural network outputs are used for that purpose, as well as how we select parameter $K$.

% to construct the top-$K$ beam pair subset $\mathcal{B}_k$.
% 

%However, the prediction confidence of the fusion pipeline varies depending upon different LoS/NLoS conditions, vehicular traffic patterns, and relative vehicle locations that may introduce obstructions. Hence, in the case of close competing classes, the algorithm must identify a larger candidate set of beam pairs to ensure that such a set contains the optimal pair. In this section, we propose an algorithm to select the cardinality of this set $K$ dynamically based on the softmax scores obtained for each prediction. $K$ becomes an important output parameter for the fusion scheme described in Sec.~\ref{sec:proposed_fusion}.

\subsection{$K$ Selection Problem Formulation}
Consider the softmax score vector  $\scores = [s_i]_{i\in\mathcal{B}} \in \mathbb{R}^{|\mathcal{B}|}$ outputed by the neural network via Eq.~\eqref{eq:fusion_network}. Recalling that $\scores$ provides a probability distribution for $(t^*,r^*)$ over $\mathcal{B}$,  the top-$K$  beam configurations Eq.~\eqref{eq:fast_p} becomes:
\begin{equation}
     \mathcal{B}_K(\scores) = 
     \underset{A\subset \mathcal{B},|A|=K}{\arg\max}~\sum_{i\in A} s_i. 
\label{eq:best_beam_pair}
\end{equation}
Hence, given scores $\scores$ and parameter $K$, $\mathcal{B}_K$ can be easily constructed by sorting $\scores$ and identifying the top-$K$ elements.
\subsection{Selecting $K$}
Parameter $K$ establishes a tradeoff between  the probability that the optimal beam pair is in $\mathcal{B}_K$ and the time it takes to determine the best (but possibly sub-optimal) beam within $\mathcal{B}_K$. This suggests selecting $K$ by optimizing an objective of the form:
$$\max_{K} \prob\left((t^*,r^*)\in \mathcal{B}_K\right) + \mu(K) $$
where $\mu:\mathbb{N}\to\mathbb{R}_+$ is a penalty increasing with the latency incurred by the choice of $K$. We discuss how to set these terms, and additional constraints we introduce, in this section.

% % ###################################Stratis
\noindent\textbf{Modeling  Probability of Inclusion.} A simple way to model the probability of the event $(t^*,r^*)\in \mathcal{B}_K$ is via the softmax scores $\scores$, as in Eq.~\eqref{eq:best_beam_pair}. We observed however that this tends to overestimate the probability of this event in practice: even if softmax scores are good for selecting the set $\mathcal{B}_K$ quickly and efficiently, a more careful approach is warranted when selecting $K$.

To that end, we leverage the empirical distribution of scores in our training set. In particular, given a score vector $\scores = [s_i]_{i\in\mathcal{B}} \in \mathbb{R}^{|\mathcal{B}|}$ and $K\in\mathbb{N}$ let 
\begin{align}
    c_K(\scores) =  \underset{A\subset \mathcal{B},|A|=K}{\max}~\sum_{i\in A} s_i 
\end{align}
be the sum of the $K$ largest scores in $\scores$. Let $I=1,\ldots,N_t$ be a sample selected uniformly at random from our training set. Let also $\scores_I$ be the corresponding softmax output layer associated with $I$, and $(t^*_I,r^*_I)\in\mathcal{B}$ the optimal pair associated with this sample. Then, given a score vector $\scores$ generated at runtime and the corresponding $\mathcal{B}_K$, we estimate the probability of the event $(t^*,r^*)\in \mathcal{B}_K$ via:
\begin{align}
p(K) =    \prob\left((t^*_I,r^*_I)\in \mathcal{B}_K(\scores_I)  \right)
\end{align}
\begin{align}
p(K;\scores) =    \prob\left((t^*_I,r^*_I)\in \mathcal{B}_K(\scores_I)  \mid c_K(\scores_I)\leq c_K(\scores)\right),\label{eq:pks}
\end{align}
where the probability  is w.r.t the random sample $I$ in the dataset. Intuitively, this captures the empirical probability that $(t^*,r^*)$ is in a random  set $\mathcal{B}_K$ constructed in the training set, conditioned on the fact that our choice of $K$ restricts these sets by bounding the quantity  $c_K$ to be at most $c_K(\scores)$. In some sense,  this allows us to link softmax scores to the variability of confidence in the construction of $\mathcal{B}_K$, itself depending upon different LOS/NLOS conditions, vehicular traffic patterns, etc., the training set is used to statistically quantify this variability.

We note that Eq.~\eqref{eq:pks} can be computed efficiently via Bayes rule, without the need to access the training set at runtime. In particular, for $c=c_K(\scores)\in \mathbb{R}_+$, $p(K;\scores)$ is equal to:
\begin{align}
\label{eq:inclusion_prob}
\resizebox{0.48\textwidth}{!}{
$\frac{\prob\left(c_K(\scores_I)\leq c\mid (t^*_I,r^*_I)\in \mathcal{B}_K(\scores_I)\right)
\prob\left((t^*_I,r^*_I)\in \mathcal{B}_K(\scores_I) \right)}{\prob\left(c_K(\scores_I)\leq c\right)}.$}
\end{align}
The constituent cumulative density functions can be computed directly  from the dataset for each $K\leq |\mathcal{B}|$, and then used at runtime.

\noindent\textbf{Incorporating Latency.}
Since the transmitter and receiver sweep all suggested beam pairs in $\mathcal{B}_K$, %including the best beam pair, $(t^*,r^*)$ is sufficient to achieve the optimum performance (i.e. beam alignment, $K\geq K^*$). Thus, we reformulate the term in Eq.~\eqref{eq:P_KeqK} by relaxing the expression such that it computes the probability of alignment ($K\geq K^*$) instead of the probability of selecting $K=K^*$. Additionally, 
we include a second term mmWave channel efficiency in the objective defined as:
\begin{equation}
    \mu (K)= \frac{T_{total} - T_{bs}^{df}(K)}{T_{total}},
\label{eq:mmwave_eff}
\end{equation}
with $T_{total}$ and $T_{bs}^{df}(K)$ being the total time for which a certain beam pair is valid and the end-to-end latency imposed by our proposed fusion based beam selection approach, respectively. We precisely analyze the end-to-end latency of our proposed beam selection approach in Sec. \ref{sec:dist_inf}. Note that the $T_{bs}^{df}$ is an increasing function of $K$. Hence, the mmWave channel efficiency is a decreasing function with respect to $K$.

\noindent\textbf{Optimization.} Combining the above terms, the final optimization problem we solve to determine $K$ given a run-time score vector $\scores$ is:
\begin{subequations}%
\label{eq:P_KgeqK}
    \begin{align}
    \underset{K}{\operatorname{max}} &\quad p(K;\scores) + \alpha \mu(K), \\
    \text{s.t.} &\quad T_{bs}^{df}(K) < T_{total},\\
                &\quad \alpha>0.
    \end{align}
\end{subequations}
In Eq.~\eqref{eq:P_KgeqK}, the first term in objective enforces the algorithm to select higher values of $K$ and ensure the alignment, when the optimum beam pair is included in the $K$ suggested beams. On the contrary, the second item  avoids selecting unnecessarily high $K$ values. The control parameter $\alpha$ in Eq. \eqref{eq:P_KgeqK} weights the importance between the two terms in the objective function.
\vspace{-2mm}

\begin{algorithm}[t!]
    \textbf{Inputs:} softmax score $\mathbf{s}$ generated by F-DL framework in Sec. \ref{sec:proposed_fusion}, $T_{total}$\;
    \textbf{Output:} $\mathcal{B}_K$\\
    % \textbf{Initialization:} Text\\
    1. Compute probability of inclusion Eq. \eqref{eq:inclusion_prob}\\ 
    2: Compute channel efficiency Eq. \eqref{eq:mmwave_eff}\\
    3: $K \leftarrow \underset{K}{\operatorname{max}} \quad p(K;\scores) + \alpha \mu(K)$\;
    4. Construct $\mathcal{B}_K$ according to Eq.~\eqref{eq:best_beam_pair} 
 \caption{Top-$K$ Beam Pair Selection}
 \label{alg:K-selection}
\end{algorithm}

\section{Dataset Description and DNN Architectures}
\label{sec:experiments}
% In this section, we introduce the publicly available Raymobtime multimodal dataset~\cite{klautau20185g} and implementation details of our proposed framework.

 {\color{black}In this section, we introduce two datasets which we use to evaluate the F-DL framework. The Raymobtime dataset~\cite{klautau20185g} is one of the widely used comprehensive multimodal dataset which has been basis of many state-of-the-art techniques. However, to give more perspective on applicability of the proposed F-DL architecture, we also collect our own ``real-world'' multimodal data, which includes real sensors, urban environment, and RF ground-truth. Further, we detail the preprocessing and implementation steps used in the proposed framework.} 

\vspace{-2mm}
\subsection{Datasets:}
\label{sec:dataset}
\subsubsection{Simulation Dataset}
\label{sec:dataset_simulation}

% \textbf{Simulation Dataset.} 
The Raymobtime multimodal dataset captures virtually with high fidelity V2X deployment in the urban canyon region of Rosslyn, Virginia for different types of traffic. A static roadside BS is placed at a height of $4$ meters, alongside moving buses, cars, and trucks.  The traffic is generated using the Simulator for Urban MObility~(SUMO) software~\cite{SUMO2018}, which allows flexibility in changing the vehicular movement patterns. The image and LiDAR sensor data are collected by Blender~\cite{blender}, a 3D computer graphics software toolkit, and Blender Sensor Simulation (BlenSor)~\cite{blensor} software, respectively. For a so called \textit{scene}, the framework designates one active receiver out of three possible vehicle types i.e. car, bus and truck. %KRC- didn't understand this... why 3? only three vehicles are moving? Also, is the vehicle Tx or RX? Fig 1 assumes former. This sentences  implies latter.\OK
For each scene, (i) the receiver vehicle collects the LiDAR point clouds and the GPS coordinates, (ii) a camera at the BS takes a picture, and (iii) the combined channel quality of different beam pairs are generated using Remcom' Wireless Insite ray-tracing software \cite{remcom}. %KRC- add citation \OK
The BS and receiver vehicle have uniform linear arrays (ULAs) with element spacing of $\lambda/2$, where $\lambda$ denotes the signal wavelength. The number of codebook elements for BS and the receiver is 32 and 8, respectively, leading to {\em 256 beam pairs}. The gap between two consecutive scenes is 30 seconds which corresponds to sampling rate of 2 samples/minute. A python orchestrator is responsible for data flow across the system to ensure the different software operations are synchronized.
% As a result, with 802.11 ad protocol, the BS and receiver needs to sweep 256 beam directions to find the best one based on equation \eqref{eq:argmax_q}. 

\begin{figure}
    \centering
    \begin{subfigure}{0.24\textwidth}
        \centering
        \includegraphics[width=\linewidth]{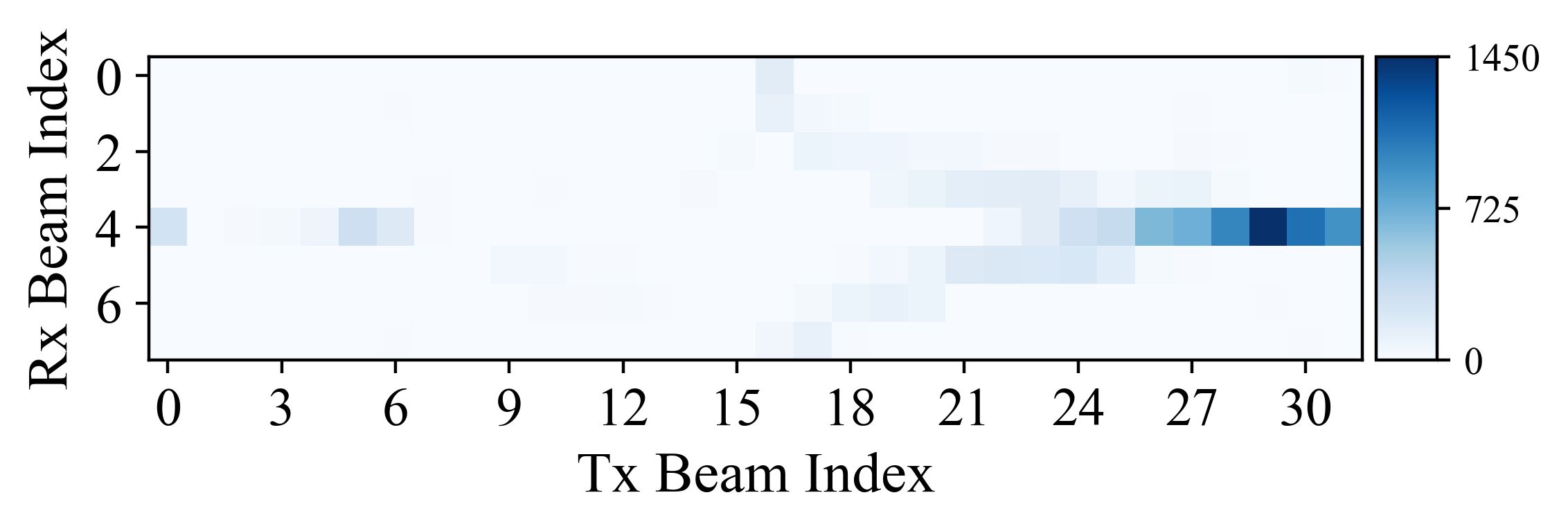}
        \caption{S008}
        \label{fig:s8_frequency}
    \end{subfigure}
    \begin{subfigure}{0.24\textwidth}
        \centering
        \includegraphics[width=\linewidth]{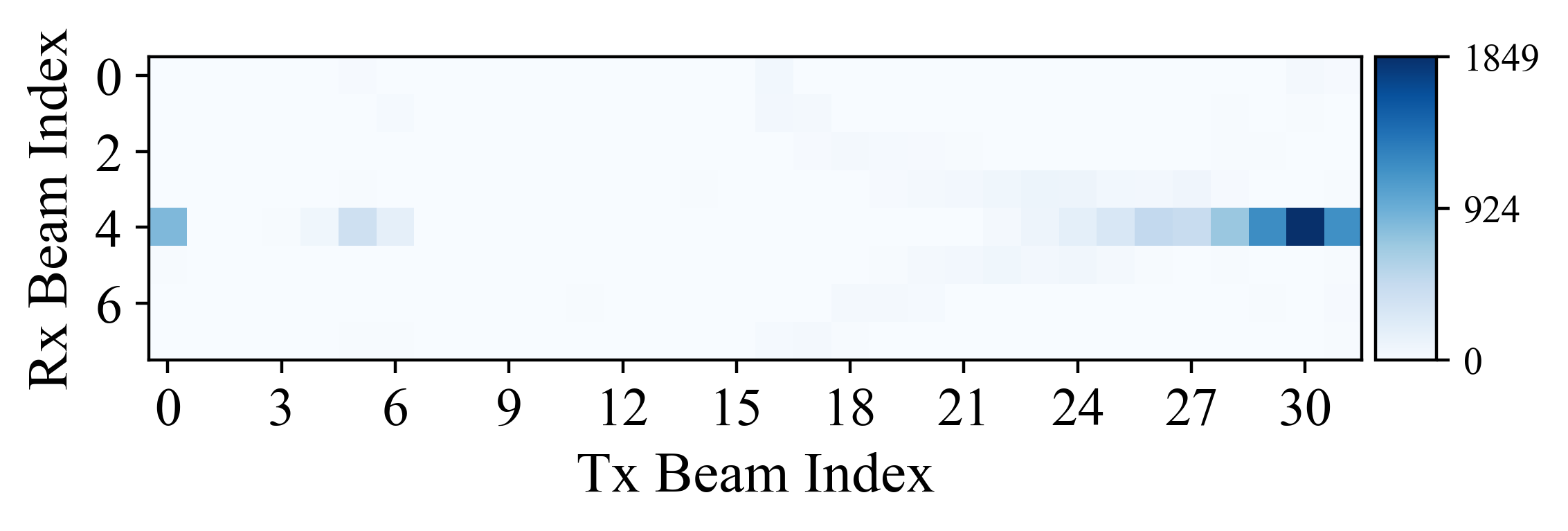}
        \caption{S009}
        \label{fig:s9_frequency}
    \end{subfigure}
    \caption{Distribution of S008 and S009 datasets.}
    \label{fig:dist_s89}
	\vspace*{-5pt}
\end{figure}

\begin{table}
    \centering
    \resizebox{0.48\textwidth}{!}{\begin{tabular}{||c|c|c|c|c||} 
    \hline
    dataset & \# of Samples  & LOS & NLOS & NLOS Percentage\\ 
    \hline\hline
     S008 & 11194 & 6482 & 4712 & 42\%\\
    \hline
     S009 & 9638 & 1473 & 8165 & 85\%\\
    \hline 
    \end{tabular}}
    \caption{Statistics of S008 and S009 datasets.}
    \label{tab:statsicts}
\end{table}
The simulation is repeated for the same scenario with two different traffic rates. We refer to these datasets as S008 and S009, which correspond to regular and rush-hour traffic, respectively. Since there are more vehicles in S009, the number of NLOS cases is higher. Tab. \ref{tab:statsicts} denotes the number of LOS and NLOS cases for both datasets. We use the S008 dataset for training and validation and S009 as the testing set. Fig. \ref{fig:dist_s89} illustrates the distribution of the classes over S008 and S009. We observe that the dataset is highly imbalanced, i.e., there is a huge variation in the number of different classes, a property that is expected due to the sparsity of mmWave links.
% F

\subsubsection{{\color{black}Real-world NEU Dataset}}
\label{sec:dataset_real}

\begin{figure}
    \centering
    \includegraphics[width=\linewidth]{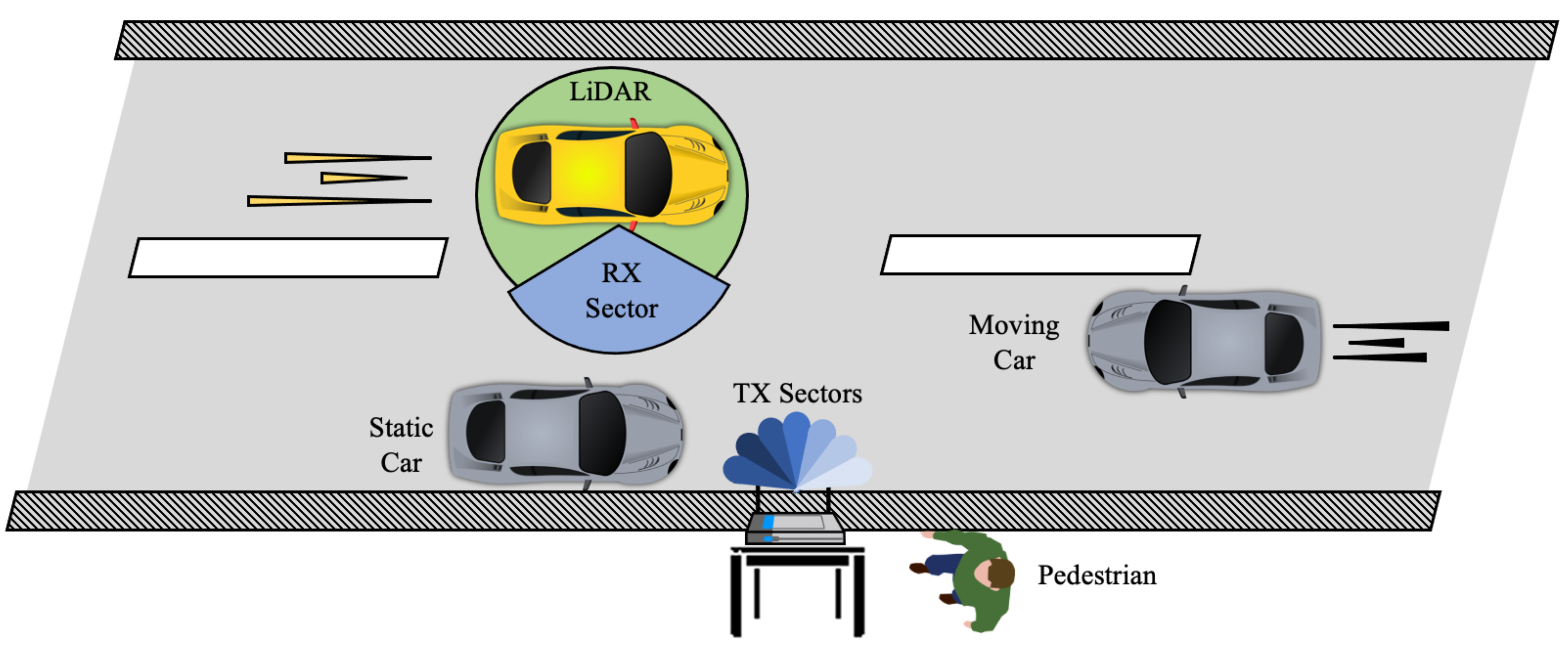}
    \caption{{\color{black}NEU dataset collection environment includes for categories as: LOS passing, NLOS by pedestrian, NLOS by static car and NLOS by moving car.}}
    \label{fig:flash_dataset_diagram}
\end{figure}

% {\color{blue}\textbf{Real-world NEU Dataset.} 
{\color{black}This dataset contains multimodal sensor observations collected in the greater metropolitan area of Boston. The experiment setting is an outdoor urban road with two-way traffic surrounded by high-rising buildings on both sides. An autonomous vehicle equipped with GPS (sampling rate 1Hz) and Velodyne LiDAR~\cite{velodyne} (sampling rate 10Hz) sensors establishes connection with a mmWave base station located at a road-side cart. The RF grand-truth is acquired using Talon AD7200 60~GHz mmWave routers with a codebook of 64 beam configurations~\cite{Steinmetzer_2017}. Each dataset sample includes the synchronized recordings of GPS and LiDAR sensors along with the grand-truth RF measurements. The data collection vehicle maintains speeds between 10-20 mph following the speed-limit of inner-city roads. The dataset setting spans a variety of four categories, including the LOS passing, blockage by pedestrian, static, and moving car with 10853 samples (116.7 GB) overall (see Tab.~\ref{tab:scenarios}). Fig.~\ref{fig:flash_dataset_diagram} denotes a diagram of the experiment setting top view. The dataset is collected during three days with different levels of humidity and weather conditions. The weather forecast information during data collection days is presented in Tab.~\ref{tab:weather}~\cite{weather}. In particular, the humidity and maximum wind speed change between 53--75\% and 8--17~mph, respectively, resulting in a rich representation of weather in the dataset and better generalization.}

{\color{black} The NEU dataset is collected to expand the feasibility study of the F-DL architecture. However, to resemble the futuristic V2X architecture, the considered framework requires tower-mounted base stations equipped with a camera. As we did not have access to such infrastructure, we collect the NEU dataset with LiDAR and GPS sensors deployed in a car. This fact does not diminish the applicability of the collected dataset, as the processed fused features from LiDAR and GPS are transmitted from car to mmWave base station following the same architecture as mentioned in Fig.~\ref{fig:intro_image}. Hence, we argue that the NEU dataset can be considered as a solid reference dataset for the beam selection task, considering the scarcity of real datasets for mmWave experiments. We pledge to release the collected real-world dataset to the community upon acceptance of this paper in our public dataset repository~\cite{genesys_dataset}. }

% {\color{red} NEED A TABLE SIMILAR (NOT SAME) TO TABLE I OF INFOCOM PAPER } \ok 

% {\color{red} NEED A TABLE WITH WEATHER CONDITIONS DAY WISE }\ok
\begin{table}
    \centering
    \resizebox{0.49\textwidth}{!}{{\color{black}\begin{tabular}{||c|c|c|c|c|c||} 
    \hline
    Day & Temperature~(F)& Humidity & {\begin{tabular}{@{}c@{}}
    Max Wind\\ Speed~(mph)
    \end{tabular}}  & {\begin{tabular}{@{}c@{}}
    Atmospheric \\Pressure~(Hg)
    \end{tabular}}  & {\begin{tabular}{@{}c@{}}
    Precipitation\\ (Inches)
    \end{tabular}}\\ 
    \hline\hline
     1 & 53-75 & 48-74\% &17 &30.13 &2.90\\
    \hline
     2 & 59-67 & 75-87\% & 13& 30 & 3 \\
    \hline
     3 & 56-68 & 54-84\% & 8& 30.37 & 3.10\\
    \hline 
    \end{tabular}}}
    \caption{{\color{black}Weather forecast on three days of data collection.}}
    \label{tab:weather}
    \vspace*{-14pt}
\end{table}

\begin{table}[t!]
    \centering
    \resizebox{0.45\textwidth}{!}{{\color{black}\begin{tabular}{||c|c|c|c||c||} 
    \hline
    Category  & Speed~(mph)  &  Scenarios & Samples\\ 
    \hline\hline
    LOS passing & 10  & --  & 1568\\
    \hline
    NLOS by pedstrain & 15  &  {\begin{tabular}{@{}c@{}}
    standing \\walk right to left \\
    walk left to right
    \end{tabular}} & 4791\\ \hline
    NLOS by static car & 15  & in front &  1506 \\ \hline
    NLOS by moving car  & 20  &  {\begin{tabular}{@{}c@{}}
    15mph same lane\\
    15mph opposite lane
    \end{tabular}} & 2988\\
    \hline 
    \end{tabular}}}
    \caption{{\color{black}Summary of different categories of NEU dataset.}}
    \label{tab:scenarios}
    \vspace*{-5pt}
\end{table}

\subsection{Preprocessing}
\subsubsection{Image}
\begin{figure}[t]
    \centering
    \begin{subfigure}{0.235\textwidth}
        \centering
        \includegraphics[width=0.91\linewidth]{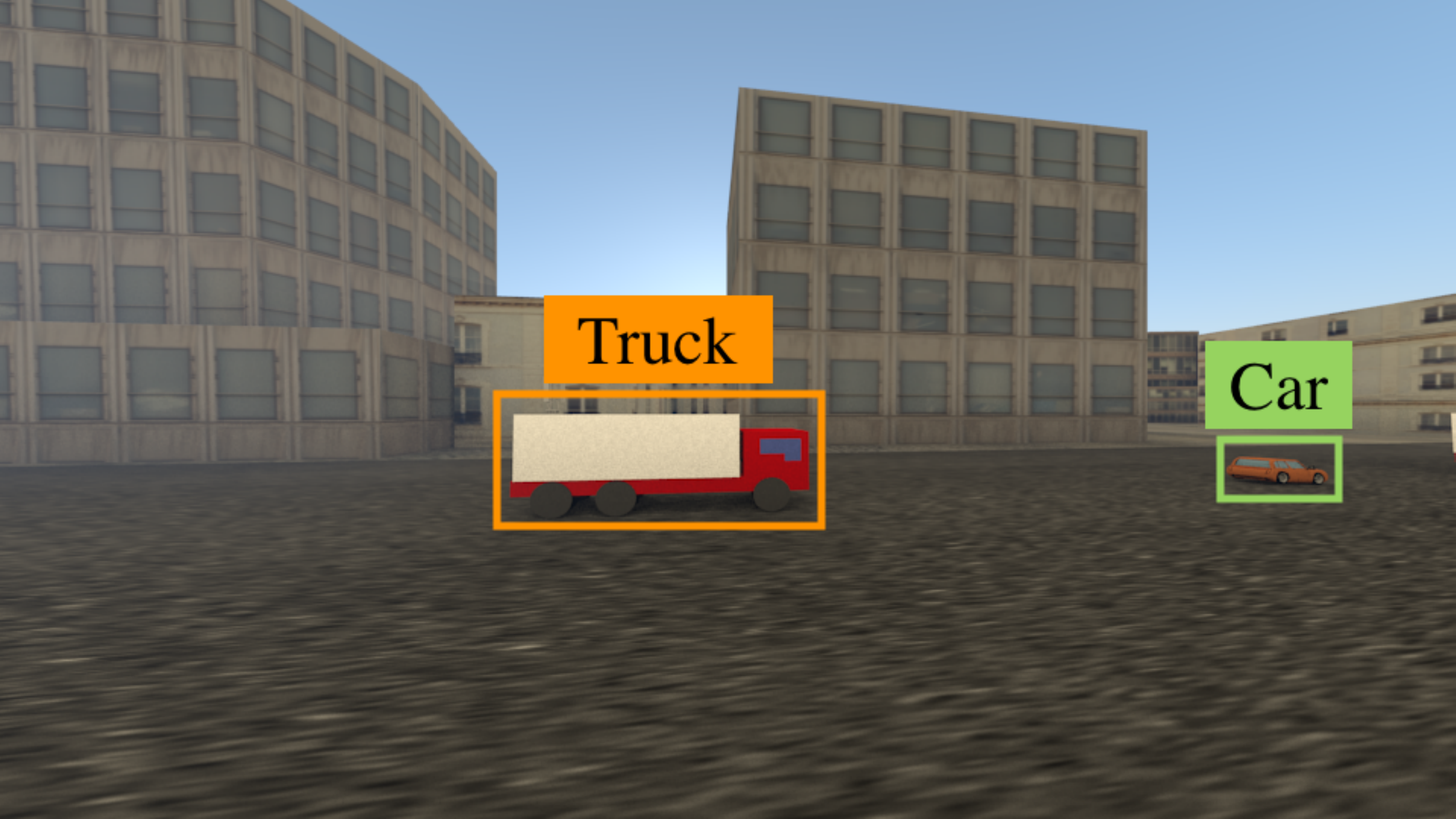}
        \caption{Raw image}
        \label{fig:input_image_preproc}
    \end{subfigure}
    \begin{subfigure}{0.22\textwidth}
        \centering
        \fbox{\includegraphics[width=0.91\linewidth]{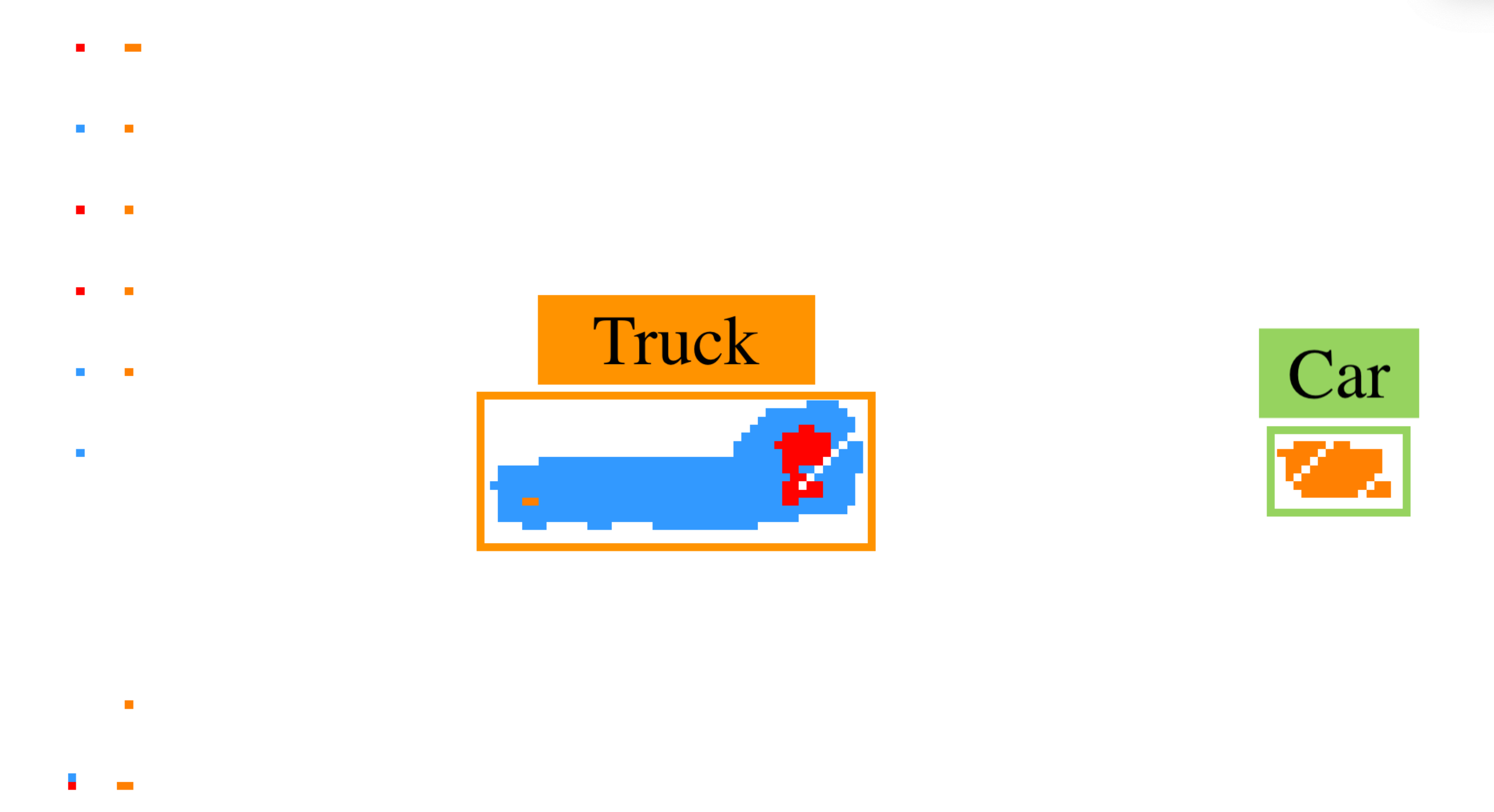}}
        \caption{Generated bit map}
        \label{fig:output_image_prepro}
    \end{subfigure}  
    \caption{An example of input and output of image preprocessing.}
    \label{fig:Result_image_feature}
    \vspace*{-10pt}
\end{figure}

To construct the dataset for the image preprocessing classifier, we manually identify and close in bounding boxes samples of bus, car, truck and background and quantize them by following the steps mentioned in Appendix  \ref{appendix:imgae_preprocessing}. We label these as background (0), bus (1), car (2), truck (3). The constructed dataset contains 22482 samples per class on average. We then train a classifier as follows. The input crops are first passed to a convolutional layer with 20 filters of kernel size (15, 15) followed by a max-pooling layer with the pool size of (3, 3) and stride size of (2, 2). The output is fed to two consecutive dense layers with 128 and 4 neurons (number of classes). Our trained classifier achieves 84\% accuracy in separating the samples of each class. In the Raymobtime dataset, the camera generates (540, 960, 3) RGB images. We empirically choose the window size of 40 and stride size 3 for our task that results in the output bit map of size (101, 185). Fig. \ref{fig:Result_image_feature} shows a sample from the dataset and the generated bit map. {\color{black}Note that the multi-object detection algorithm can be easily extended to any type of vehicle by including the samples from new vehicles in the training set~\cite{felzenszwalb2004efficient,cheng2001color}}. We evaluate the delay cost of image preprocessing in Sec. \ref{sec:delay_preproc}.

\subsubsection{LiDAR}
The maximum distance for LiDAR is set to 100 meters in the Raymobtime dataset, and the zone of space is limited in each axis as, $(X_{\min}, X_{\max})=(744, 767)$, $(Y_{\min}, Y_{\max})=(429, 679)$, $(Z_{\min}, Z_{\max})=(0, 10)$, where the static base station is located at $[746, 560, 4]$ within this Cartesian coordinate system. Moreover, the histogram bin size along the three spatial dimensions is set as $(1.15, 1.25, 1)$, respectively. By following the steps mentioned in Sec. \ref{sec:preprocessing_lidar}, we generate a compact $(20, 200, 10)$ representation of the environment where the BS, target vehicle, and obstacles are identified and marked with different indicators. {\color{black} For NEU dataset, we use the maximum LiDAR distance of 80 meters and map the LiDAR point clouds to a compact $(20,20,20)$ representation in each axis.}
% ('Quantization paramethers', {'Zmax': 10, 'Ymax': 205.00021731824822, 'Xmax': 33.444200263353, 'Xmin': -221.98978455647742, 'Yp': 13.021759363161953, 'Xp': 12.77169924099152, 'Ymin': -55.434969944990875, 'Zp': 0.5, 'Zmin': 0})
% The section and subsection headings should follow the same sequence as the proposed method.

\vspace{-2mm}
\subsection{Implementation Details}
\begin{figure}
    % https://app.diagrams.net/#G1ART4fli7Ww_L1Rk5DnZlDGS55IP0J4_A
    \centering
    \begin{subfigure}{0.095\textwidth}
        \centering
        \includegraphics[width=\linewidth]{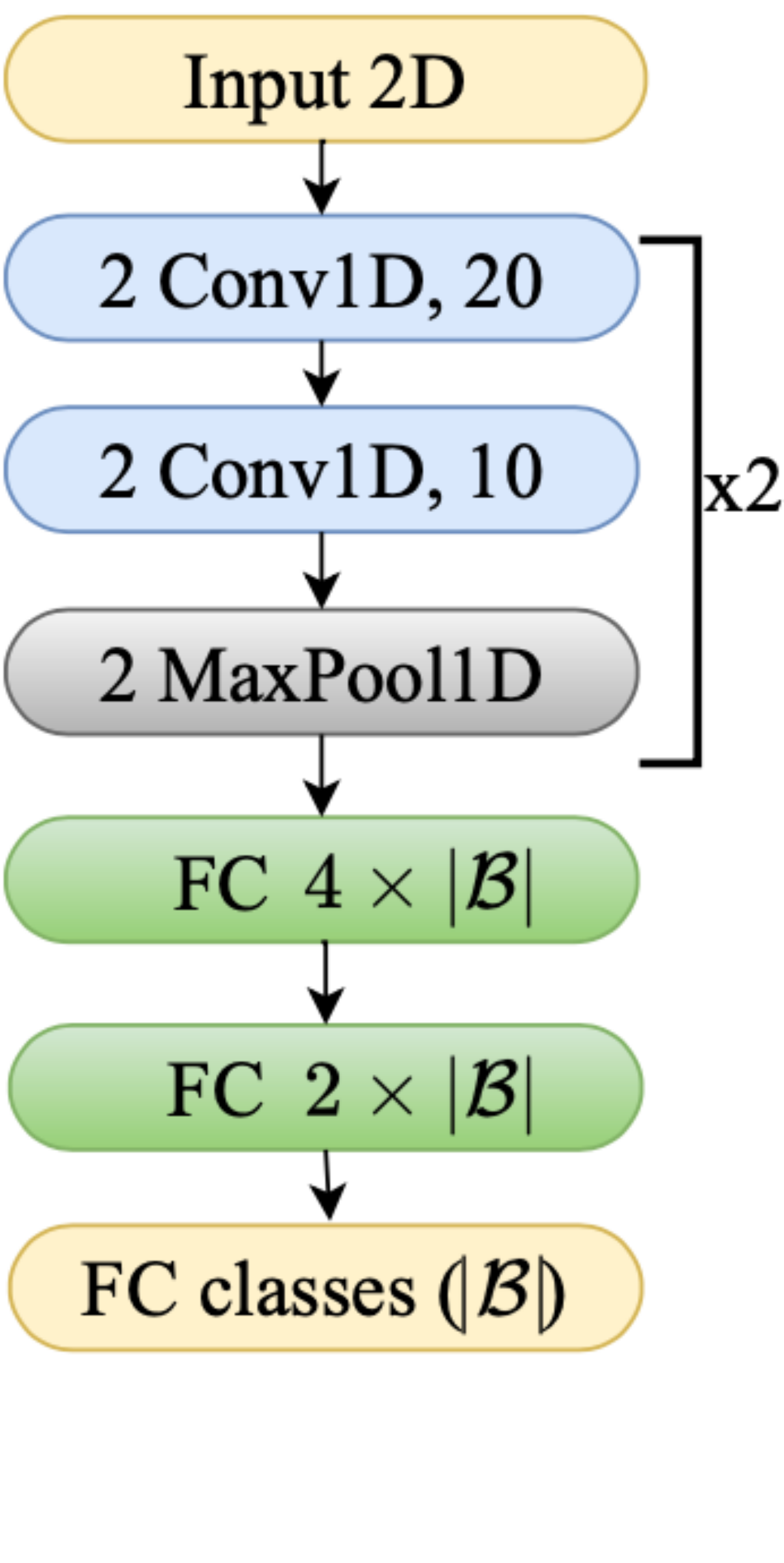}
        \caption{~GPS}
        \label{fig:model_coord}
    \end{subfigure}
    \begin{subfigure}{0.095\textwidth}
        \centering
        \includegraphics[width=\linewidth]{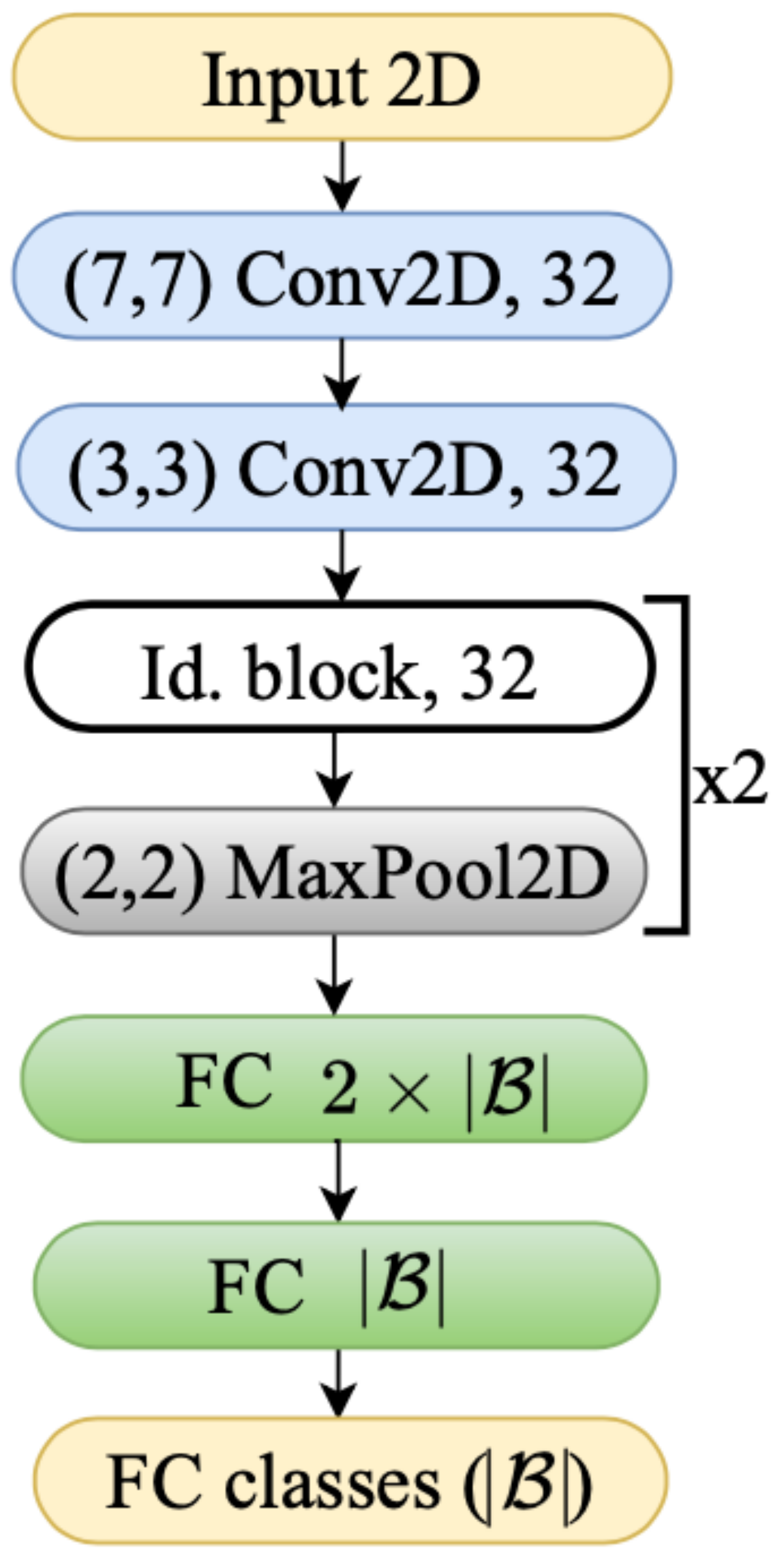}
        \caption{Image}
        \label{fig:model_Img}
    \end{subfigure}  
    \begin{subfigure}{0.095\textwidth}
      \centering
      \includegraphics[width=\linewidth]{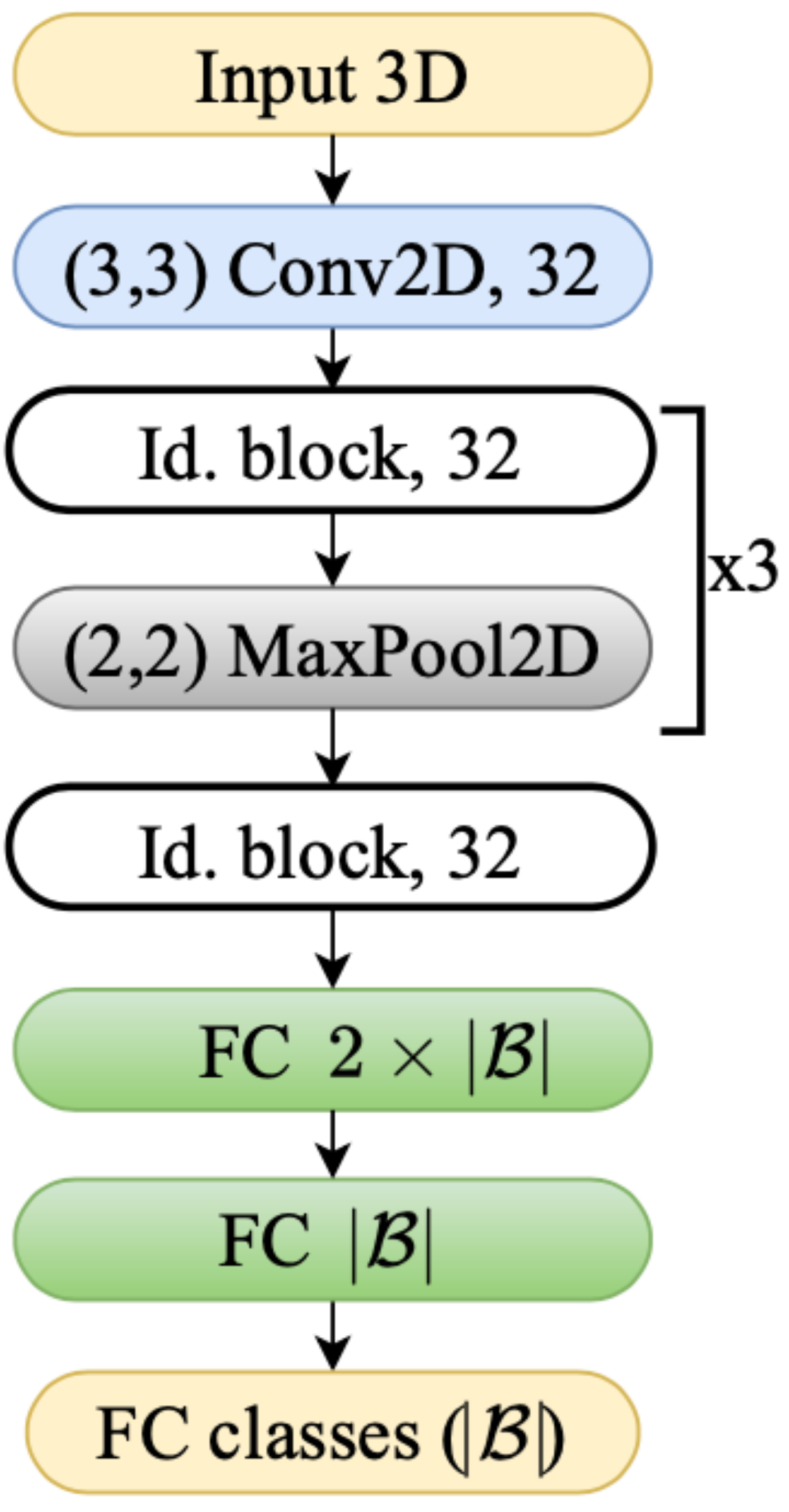}
        \caption{LiDAR}
        \label{fig:LIDAR_model}
    \end{subfigure}
    \begin{subfigure}{0.095\textwidth}
      \centering
      \includegraphics[width=\linewidth]{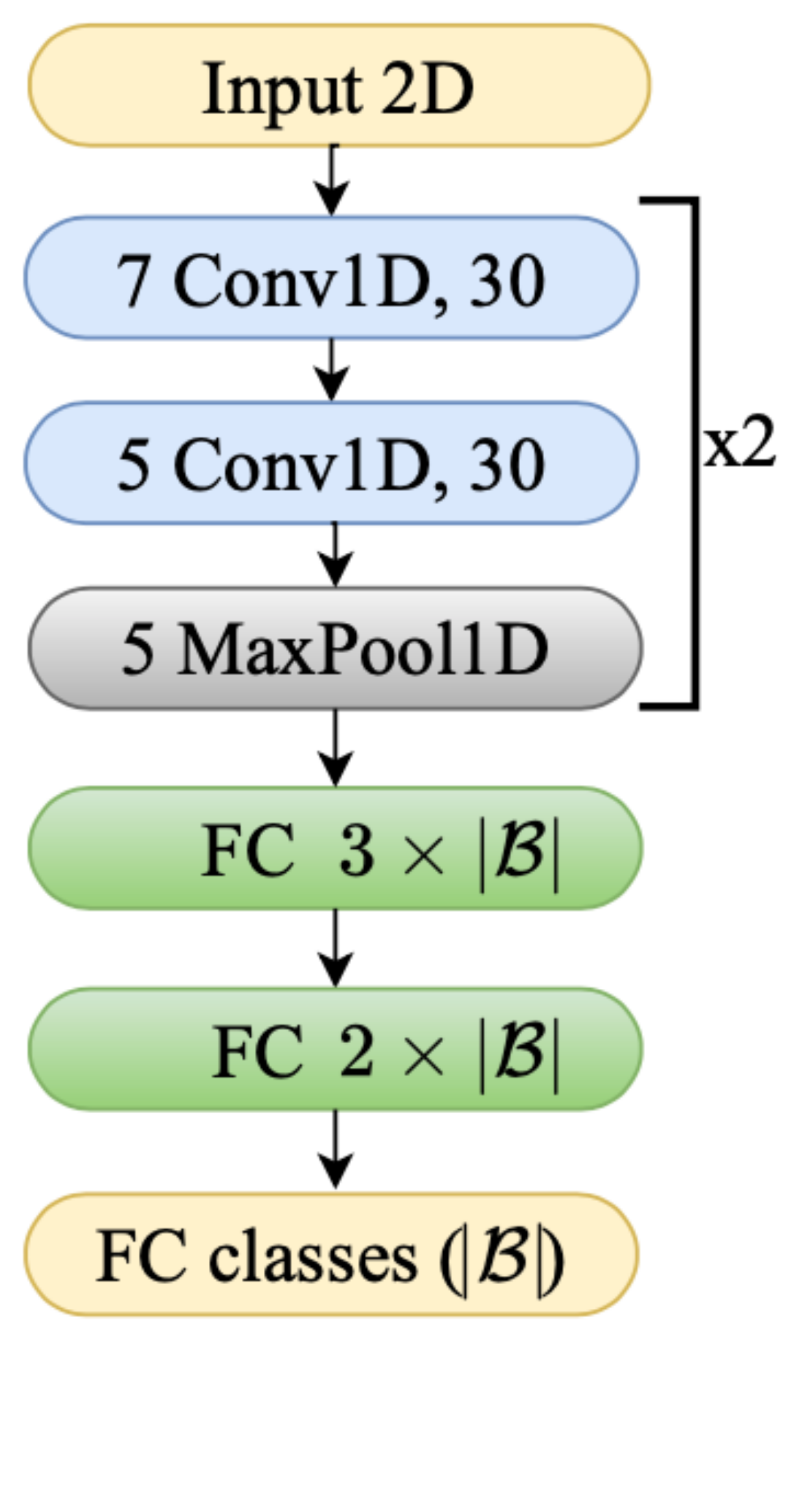}
        \caption{Fusion}
        \label{fig:fusion}
    \end{subfigure}
    \begin{subfigure}{0.079\textwidth}
      \centering
      \includegraphics[width=\linewidth]{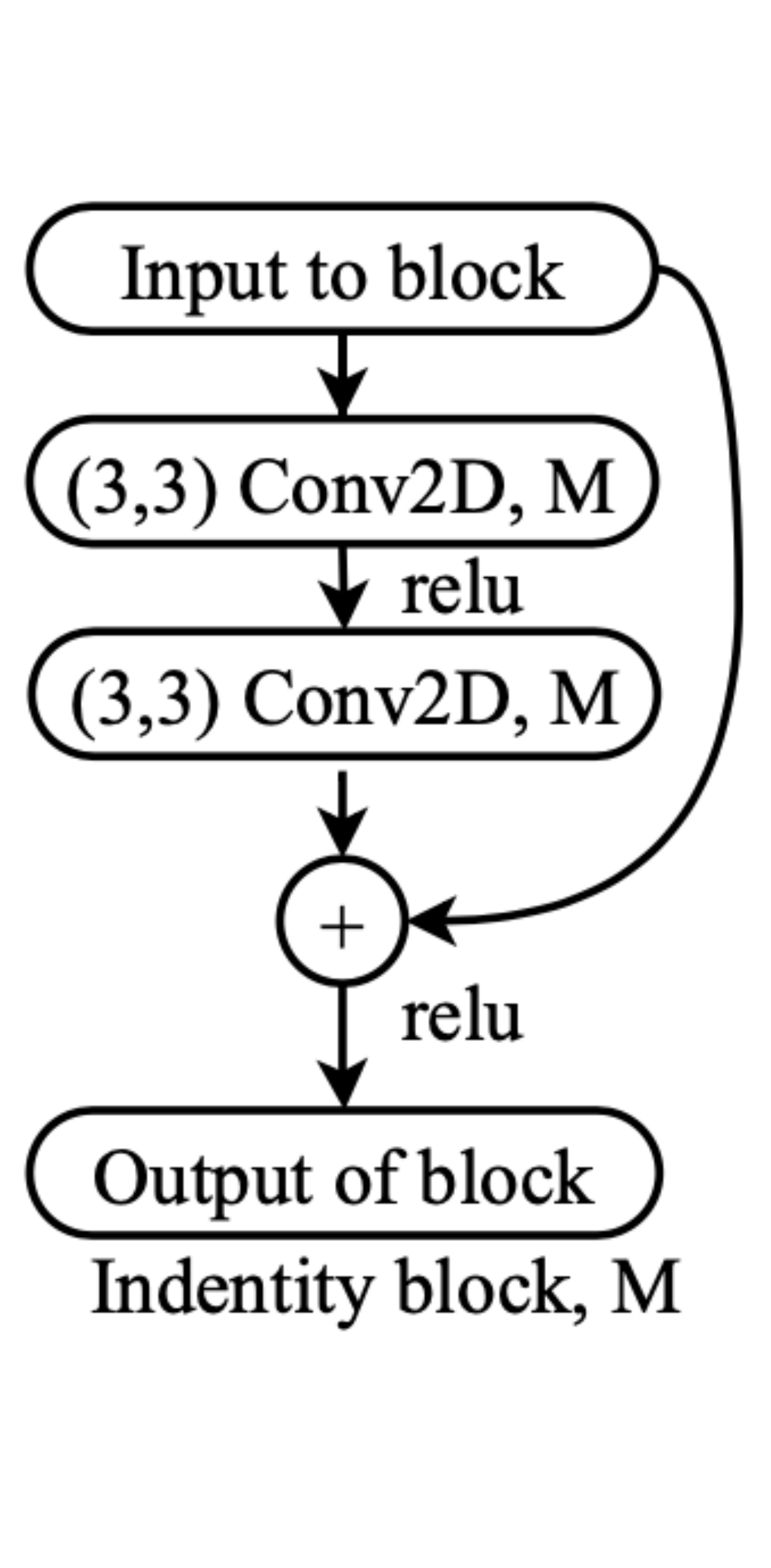}
        \vspace*{0.01pt}
        \caption{Identity}
        \label{fig:idenetity block}
    \end{subfigure}
    \caption{Proposed architectures for unimodal and fusion networks.}
    \label{fig:model_architectures}
    \vspace*{-10pt}
\end{figure}

% \subsubsection{Designing Base Single-modal Networks}
Our proposed fusion pipeline consists of three unimodal networks per modality followed by a fusion network as presented in Fig.~\ref{fig:fusion_pipeline}. We first design each unimodal network tuned to {\color{black} each} dataset which takes either raw (for coordinate) or preprocessed (LiDAR and image) data as input and generate the latent embeddings to be fed to the fusion network.
% For all models, we use convolutional modules followed by dense modules.  
For GPS unimodal network, we design a model that uses 1-D convolutional layers (see Fig.~\ref{fig:model_coord}). This enables capturing the correlation between the latitude and longitude, simultaneously. {\color{black}Our custom designed model for the preprocessed images (see Sec.~\ref{sec:preprocessing_img}) is inspired by ResNet~\cite{resnet} that uses identity connections to avoid the gradient vanishing problem commonly seen in deep architectures, by creating a direct path for the gradient during backpropagation.} Each such \textit{identity block} contains 2 convolutional layers %with 32 kernels of size $3\times3$
and an identity shortcut that skips these 2 layers, followed by a %$2\times2$
max-pooling layer, as shown in Fig. \ref{fig:idenetity block}. For LiDAR input, we also design a model structure similar to ResNet (see Fig.~\ref{fig:LIDAR_model}).
%The size of filters in our custom base single-modal networks is precisely adjusted based  on input data. 
Note that while the input to image and LiDAR models are 2D and 3D, the majority of elements are zero due to filtering the irrelevant data during preprocessing. We also use max-pooling layers after convolutional layers for feature down-sampling and dropout of 0.25 after fully-connected layers to avoid overfitting.

% We set the dimensionality of each single-modal network (latent embedding generator) to be equal to number of beam pairs, i.e., $d = 256$.
{\color{black} The representation capacity of each network including latent embedding generators scales with the number of classes $|\mathcal{B}|$ in each dataset, 256 and 64 for Raymobtime and NEU, respectively.}
Though increasing the number of neurons generally improves the representation capacity of base unimodal architectures, we find {\color{black} having} neurons {\color{black}equal to the number of classes} to be sufficient for our task. We design a fusion network as depicted in Fig.~\ref{fig:fusion} that takes as input the concatenated latent embedding of each modality. Ultimately, the last dense layer with the number of classes outputs the predicted score of each beam pair. For all models, we exploit categorical cross-entropy loss with batch size of 32 and training epochs of 100 {\color{black}and 400 for Raymobtime and NEU dataset with an earlier stopping point of patience 10. Moreover, we apply $\ell_1$ and $\ell_2$ kernel regularizers on dense layers with parameters $10^{-5}$ and $10^{-4}$, respectively}. We use Adam~\cite{adam} as optimizer with $\beta=(0.9,0.999)$ and initialize the learning rate to 0.0001. We tuned all the hyperparameters on a validation set via holding out 17\% {\color{black}and 10\% of the training data for each dataset.} %(S008).
% Moreover, we choose rectified linear units (ReLU) as activation function for convolution layers. The activation functions introduce a non-linearity on feature maps that allows models to learn a wide variety of functions. Compared to other activation functions, ReLU sets negative values to zero, which results in sparser outputs. The sparsity may be pursued for multiple reasons in a CNN but it mainly provides robustness to small changes in input. 

% image and lidar con2D
% kernel size 
% last layer

\vspace{-2mm}
\section{End-to-End Latency Analysis with Distributed Inference}
\label{sec:dist_inf}
In this section, we explore the important design details and performance trade-offs related to centralized/distributed inference. Moreover, we answer the following question: \textit{What is the end-to-end latency of beam selection with our proposed method?} %Note that the end-to-end system includes the following five steps, as described in Sec. \ref{sec:system_des}. 
% (i) data collection and preprocessing (ii) sharing the extracted features from vehicle to MEC (iii)  inference/sharing selected beams from MEC to vehicle (iv) standards-defined actual beam-sweeping time for the suggested $K$ beam configurations, as described in section \ref{sec:system_des}. 
% The NVIDIA V100 GPU with 32GB memory is used to evaluate the execution time of our proposed method.

\vspace{-2mm}
\subsection{Data Collection and Preprocessing}
\label{sec:delay_preproc}
%With the recent development in using advanced processing units in the sensors, we assume that the data collection happens instantaneously~\cite{cecchinel2014architecture}.
% {\color{blue}DR: if possible give reference}. \ok
%For instance, the 
Current LiDAR sensors support pulse rate, i.e., the number of discrete laser ``shots” per second that the LiDAR is firing, of 50,000 to 150,000 pulses per second, while 35~cm precision can be achieved with 8~pulses/$m^2$~\cite{carter2012introduction}.
% {\color{blue}DR: must give reference}. \ok
The GPS sensor data does not require any preprocessing and the LiDAR preprocessing has a negligible latency that can be further reduced by exploiting parallel processing. For image sensor data, we measure the delay of our proposed object detection algorithm described in Appendix. \ref{appendix:imgae_preprocessing} by passing a single sample 100 times and calculating the average required time for generating bit maps. Accordingly, our proposed image preprocessing pipeline generates the bit maps in $1.30\,ms$ on average. As a result, our preprocessing pipeline runs in 1.30~ms on average ($T_{\mathrm{process}}=1.30\,ms$). {\color{black} Note that image preprocessing is applied on Raymobtime dataset only.}

\subsection{Sharing Features between Vehicle and MEC}
Data collected at vehicular locations can incur different relaying costs to the MEC, depending upon the sensor modality. GPS coordinates, both latitude and longitude, can be expressed in 6 Bytes, while the raw LiDAR point cloud requires $\sim$1-1.5~MBytes for complete transfer. % KRc- collected in how much time? There must be some indication what one collection round is? 
%Hence, sending the raw LiDAR and coordinates comes with a huge transmission cost that is in contrary to our goal, i.e. reducing the beam selection time. One possible approach is to 
One  possible  approach  is  to  relay  the  GPS measurements \textit{as is} while subjecting the LiDAR data to additional preprocessing step as discussed in Sec.~\ref{sec:preprocessing_lidar}. This step maps the raw LiDAR point clouds to a ridge representation with size (20, 200, 10) that can be shown with $\sim$320 KBytes (78\% less than raw LiDAR point clouds) {\color{black} for Raymobtime dataset. Using the aforementioned prepossessing reduces the data from 0.9~MByte to 64~KByte for NEU dataset as well.} We can further improve the data transmission speed from vehicle to the MEC by sending the fused high level latent embeddings of LiDAR and GPS. Recall that we extract this information at an intermediate layer of the neural network (see Sec. \ref{sec:fusion_network}). With our proposed distributed inference design, the raw coordinates and LiDAR data is translated to an array with $2\times|\mathcal{B}|$ elements that is expressed with only $\sim$4 KBytes {\color{black} and $\sim$1~KBytes for Raymobtime and NEU datasets, respectively} ($\sim99\%$ reduction in size than raw data), which is even more compressed and requires less bandwidth within the sub-6 GHz control channel.
% {\color{blue}control channel (DR: SHOULD IT NOT BE SUB-6 GHZ DATA CHANNEL?)} \Yes \ok

%As mentioned in Sec.~\ref{sec:system_des}, we rely on existing standards of either IEEE 802.11p standard for wireless access in vehicular environments or Long-Term Evolution (LTE) for sub-6 GHz data and control channels. 
Tab. \ref{tab:backchannelTime} illustrates the number of bytes and the minimum/maximum experienced delay while transmitting the compressed extracted features of coordinate and LiDAR over the sub-6~GHz data channel. The achievable throughput is assumed to be 3-27~Mbs and 4.4-75~Mbs for 802.11p~\cite{wang2011ieee} and single input single output~(SISO) LTE~\cite{sesia2011lte}, respectively. 

% http://www.techplayon.com/lte-fdd-system-capacity-and-throughput-calculation/
% , where the achievable throughput assumed to be 3-27Mbs according to 802.11p standard for wireless access in vehicular environments \cite{wang2011ieee}. 
Additionally, the fused features are difficult to interpret by third parties and provide a level of abstraction to the raw data. From Tab. \ref{tab:backchannelTime}, we observe that the data channel delay reduces drastically with the distributed inference. Without loss of generality, we use the maximum imposed delay of control signaling from vehicle to MEC being $(T_{\mathrm{data}}= 1.332\,ms)$ {\color{black}for Raymobtime and $(T_{\mathrm{data}}= 0.33\,ms)$ for NEU datasets} to calculate the overall end-to-end latency.
% F
% \begin{table}
%     \centering
%     \resizebox{0.48\textwidth}{!}{\begin{tabular}{||c|c|c|c|c|c|c||} 
%     \hline
%     \multirow{2}{1cm}{Method} & \multirow{2}{1cm}{\# Bytes} & \multicolumn{2}{c|}{Min Reqd. time ($\,ms$)} & \multicolumn{2}{c|}{Max Reqd. time ($\,ms$)}\\ 
%     \cline{3-6}
%      &  & 802.11p & LTE & 802.11p & LTE\\
%     \hline\hline
%     Raw data & 1.506 MB & 55.77  & 20.08  & 502  & 342.27  \\
%     \hline
%      Preprocessed & 326 KB  & 12.07  & 4.34  & 108.66  & 74.09 \\
%     \hline 
%     {\begin{tabular}{@{}c@{}}
%     High-level \\fused features
%     \end{tabular}} & 4 KB  & 0.148  & 0.053  & \textbf{1.332 } & 0.90 \\
%     \hline
%     \end{tabular}}
%     \caption{The required time for sharing the data with MEC with three data sharing strategies.}
%     \vspace*{-10pt}
%     \label{tab:backchannelTime}
% \end{table}

\begin{table*}
    \centering
    \resizebox{0.9\textwidth}{!}{\begin{tabular}{||c||c|c|c|c|c||c|c|c|c|c||} 
    \hline 
    & \multicolumn{5}{c||}{Raymobtime} &\multicolumn{5}{c||}{{\color{black}NEU}}\\
    \cline{2-11}
    \multirow{3}{1cm}{Method} & \multirow{3}{1cm}{\# Bytes} & \multicolumn{2}{c|}{Min Reqd. time ($\,ms$)} & \multicolumn{2}{c||}{Max Reqd. time ($\,ms$)} & \multirow{3}{1cm}{{\color{black}\# Bytes}} & \multicolumn{2}{c|}{{\color{black}Min Reqd. time ($\,ms$)}} & \multicolumn{2}{c||}{{\color{black}Max Reqd. time ($\,ms$)}}\\ 
    \cline{3-6} \cline{8-11}
     &  & 802.11p & LTE & 802.11p & LTE & & {\color{black}802.11p} & {\color{black}LTE} & {\color{black}802.11p} & {\color{black}LTE}\\
    \hline
    % Raw data & 1.506 MB & 55.77  & 20.08  & 502  & 342.27 & {\color{blue}0.9 MB} & {\color{blue}33.32}  & {\color{blue}12.02}  & {\color{blue}300.60}  & {\color{blue}204.54} \\
    \hline
     Preprocessed & 326 KB  & 12.07  & 4.34  & 108.66  & 74.09 & {\color{black}64 KB}  & {\color{black}2.37}  & {\color{black}0.85}  & {\color{black}21.33}  & {\color{black}14.55}\\
    \hline 
    {\begin{tabular}{@{}c@{}}
    High-level \\fused features
    \end{tabular}} & 4 KB  & 0.148  & 0.053  & \textbf{1.332 } & 0.90 & {\color{black}1 KB}  & {\color{black}0.037}  & {\color{black}0.013}  & {\color{black}\textbf{0.33}} & {\color{black}0.225}\\
    \hline
    \end{tabular}}
    \caption{The required time for sharing the data with MEC $(T_{\mathrm{data}})$ for three data sharing strategies {\color{black} for Raymobtime and NEU datasets.}}
    % {\color{red}CHANGE COLOR TO BLUE FOR EACH THE TEXTS IN THE TABLE WHICH AS BEEN NEWLY ADDED}}\ok
    \vspace*{-10pt}
    \label{tab:backchannelTime}
\end{table*}

\subsection{Inference and Sharing Selected Beams with Vehicle}
In order to evaluate the inference delay, we pass input data, i.e., the latent embedding of all modalities, through our pipeline and measure the prediction time by setting a timer and subtracting the timestamp before and after prediction. We note that the average inference time of our proposed fusion approach is $0.37\,ms$. On the other hand, sending the selected $K$ beams from MEC to vehicle over the sub-6 GHz control channel requires at most 2KB (256 elements) {\color{black} and 0.5KB (64 elements) for Raymobtime and NEU datasets, respectively. That takes $0.66\,ms$ {\color{black}and $0.16\,ms$ as maximum required time}, and results in a cumulative delay ($T_{\mathrm{control}}$) of $1.03\,ms$ and $0.53\,ms$ for each dataset, respectively. Similar to the previous section, we consider the highest imposed delay related to using IEEE 802.11p standard as our reference.}
% Similar to the previous section, we consider the highest imposed delay $1.03\,ms$ related to using IEEE 802.11p standard as our reference ($T_{control}= 1.03\,ms$). %KRC- we don't say what the MEC is for us? So how does the processing time generalize or how is the reader going to place things in context?

\subsection{Impact on Beam-sweeping Latency: Case Study in 5G-NR}
% To calculate the beam sweeping time, we first discuss the time requirement of exhaustive beam search in 5G-NR standard. Next, we show improvement in beam sweeping time using the proposed approach even after following the same norms of 5G-NR standard.
We first discuss the time requirement of exhaustive beam search in 5G-NR standard. Next, we calculate the required time for sweeping only the selected $K$ beam pairs by following the same norms as 5G-NR standard.
\subsubsection{Beam Selection Latency in 5G-NR}
\label{subsec:beamTime}
For evaluating a 5G-NR standard compliant beam selection process in the mmWave band, we consider a transmitter-receiver pair with the codebook sizes $M$ and $N$, respectively. With analog beamforming, we have a total of $|\mathcal{B}|= M \times N$ combinations (see  Sec.~\ref{sec:systemModel}). During the initial access, the gNodeB and user exchange a number of messages to find the best beam pair. In particular, the gNodeB sequentially transmits synchronization signals~(SS) in each codebook element $t_m \in C_{Tx}$. Meanwhile, the receiver also tunes its array to receive in different codebook elements $r_n\in C_{Rx}$ until all possible beam configurations are swept. The SS transmitted in a certain beam configuration is referred as the SS block, with multiple SS blocks from different beam configurations  grouped into one SS burst. The NR standard defines that the SS burst duration ($T_{ssb}$) is fixed to $5\,ms$, which is transmitted with a periodicity ($T_p$) of $20\,ms$~\cite{barati2020energy}. In the mmWave band, a maximum of 32 SS blocks fit within a SS burst, which allows for 32 different beam pairs to be explored within one SS burst. Hence, in order to explore all beam pair combinations, a total of $|\mathcal{B}|$ SS blocks are required to be transmitted. Given the limit on SS blocks within a SS burst, the total time to explore all beam pairs ($T_{bs}^{nr}$) can be expressed as:
\begin{equation}
    T_{bs}^{nr}(|\mathcal{B}|) = T_p \times \left \lfloor\frac{ |\mathcal{B}| - 1}{32} \right \rfloor + T_{ssb},
    \label{eq:T_nr}
\end{equation}
where $T_p=20\,ms$ and $T_{ssb}=5\,ms$ correspond to periodicity and SS burst duration, respectively. Note that if a certain number of beam pairs are not explored within the first SS burst ($|\mathcal{B}| > 32$), there is an increasing delay given the separation $T_p$ between SS bursts. On the other hand, exploring a number of pairs smaller than 32 will introduce the same overhead as if a total of 32 options were searched, given that $T_{ssb}$ has a fixed duration of $5\,ms$. Similarly, this can be extended to any number $|\mathcal{B}|$ that is not a multiple of 32.

\subsubsection{Improvement in Latency through Proposed Approach}
Our proposed approach reduces the beam search space from $|\mathcal{B}|$ to a subset of $K\ll |\mathcal{B}|$ most likely beam candidates, derived from  Algorithm~\ref{alg:K-selection}. We recall that the NR standard  assumes that up to 32 can be swept within $5\,ms$. Thus, we define the time to explore one single beam as $T_b=5\,ms/32=156\,ns$. Then, the required time for sweeping the selected top-$K$ beam pairs can be expressed as:
\begin{equation}
    T_{\mathrm{sweep}}(K) =  T_p \left \lfloor\frac{K-1}{32}\right \rfloor + T_b ~(1+(K-1)\bmod{32}).
    \label{eq:T_sweep}
\end{equation}
\subsection{End-to-end Latency Calculation}
Considering the aforementioned four steps, the overall beam selection overhead following our proposed data fusion approach ($T_{bs}^{df}$) with distributed inference is expressed as:
\begin{equation}
    \begin{split}
        &T_{bs}^{df}(K) = T_{\mathrm{process}} +T_{\mathrm{data}} +T_{\mathrm{control}}+T_{\mathrm{sweep}}(K),
    \end{split}
    \label{eq:T_df}
\end{equation}
where the first three terms can be approximated by $3.662\,ms$ {\color{black} and $0.86\,ms$ for Raymobtime and NEU datasets, respectively}. Note that the distributed inference play a pivotal role in reducing the overhead associated with sharing the situational state of the vehicle with the MEC ($T_{\mathrm{data}}$). We validate the improvement in overall beam selection time using the proposed distributed inference (Eq.~\ref{eq:T_df}) approach rather than the traditional brute-force approach offered by the state-of-the-art 5G-NR (Eq.~\ref{eq:T_nr}) standard in Sec.~\ref{subsec:beam_selection_speed}.

% The improvement in overall beam selection time using the proposed distributed inference (Eq.~\ref{eq:T_df}), than the traditional centralized 5G NR (Eq.~\ref{eq:T_nr}), is presented in Sec.~\ref{subsec:beam_selection_speed} for the experiment described in Sec.~\ref{sec:experiments}.

%%%%%%%%%%%%%%%%%%%%%%%%%%%%%%%%%%%%%%%%%%%%%%%%%%%%%%%%%%%%
%%%%%%%%%%%%%%%%%%%%%%%%%Extras%%%%%%%%%%%%%%%%%%%%%%%%%%%%%
%%%%%%%%%%%%%%%%%%%%%%%%%%%%%%%%%%%%%%%%%%%%%%%%%%%%%%%%%%%%
% where $T_{ds}= 1.332ms$ is the data sharing time (Table.~\ref{tab:backchannelTime}), $T_{nn}= 1.67ms$ is the NN inference time and $T_{ks}=0.22ms$ is the time taken to communicate the \textit{to-be-swept} $K$ beam-pairs from the MEC to the vehicle.

%  Systems used in 2012 were capable of up to 300,000 pulses per second. More commonly, the data are captured at approximately 50,000 to 150,000 pulses per second.

% Image feature extractor takes 1.30 ms
% inference time (only CNN) = 0.37 ms

% From V to I = 1.332ms 

% From I to V depends on the number of K's that you want to transmit. At most with 256 elements(2KB) it will be  0.66ms. 

% Specifically in this work, we partition the 3-D space into fixed-sized cuboid regions, and quantize information by replacing individual LiDAR point locations with the indices of the cubes in which they occur (we describe this later in Sec. \ref{sec:preprocessing}).
\section{Results and Discussions}
\label{sec:performance}
\begin{table*}[t]
    \setlength{\tabcolsep}{5pt}
    \centering
\resizebox{1.0\textwidth}{!}{
     \begin{tabular}{|c|c|c|c|c|c|c|c|c|c|c||} 
     \hline 
    Modalities & Top-$1$ & Top-$2$ & Top-$5$ &Top-$10$ & Top-$25$ & Top-$50$ & Weighted & Weighted & Weighted & KL\\
     & Accuracy  & Accuracy &Accuracy  & Accuracy & Accuracy & Accuracy & Recall & Precision & F1 score & divergence \\
     \hline \hline
    Coordinates & 12.32\% & 31.51\% & 55.61\% & 77.93\% & 88.5\% & 95.14\% & 2\% & 12\% & 3\% & 3.02\\ 
     \hline 
    Image & 12.39\% & 26.84\% & 55.38\% & 71.65\% & 88.05\% & 95.01\% & 7\% & 12\% & 3\% & 2.9051 \\
     \hline
    LiDAR & 46.23\% & 64.67\%& 82.43\% & 89.95\% & 96.11\% & 98.13\% & 47\% & 46\% & 45\% & 0.1738\\
     \hline
    Coordinates, Image & 25.76\% & 44.88\% & 74.18\% & 86.29\% & 94.78\% & 97.89\% & 21\% & 26\% & 22\% & 0.5432\\
     \hline
    Coordinates, LiDAR& 55.42\% & 74.54\% & 85.51\% & 91.41\% & 96.75\% & 98.56\% & 55\% & 55\% & 54\% & 0.1357 \\
    \hline
    Image, LiDAR & 54.52\% & 73.08\% & 84.83\% & 91.23\% & 96.78\% & 98.50\% & 55\% & 55\% & 54\% & 0.1428\\
     \hline
    Coordinate, Image, LiDAR & \textbf{56.22\%} & \textbf{74.08\%} & \textbf{85.53\%} & \textbf{91.11\%} & \textbf{96.56\%} & \textbf{98.60\%} & \textbf{55\% }& \textbf{56\%} & \textbf{55\%}& \textbf{0.1314}\\ 
     \hline
    \end{tabular}}
        \caption{Performance of proposed unimodal and fusion when trained on S008 and tested on S009 Raymobtime dataset.}
    \label{tab:results}
    \vspace*{-10pt}
\end{table*}

In this section, we provide the results of our proposed method using the datasets described in Sec. \ref{sec:dataset}. We use Keras 2.1.6 with Tensorflow backend (version 1.9.0) to implement our proposed beam selection approach. The source codes for our implementation are available in~\cite{githubcodes}. To judge the efficiency of proposed beam selection approach on multi-class, highly-imbalanced, multimodal Raymobtime~\cite{klautau20185g} {\color{black}and NEU} datasets, we use four evaluation metrics that capture the performance from different aspects, including top-$K$ accuracy, weighted F-1 score, KL divergence and throughput ratio. We provide the detailed definitions of these metrics in Appendix~\ref{app:evalution_metrics}. {\color{black} We first analyze the performance of proposed fusion deep learning method on Raymobtime dataset, and then further justify the improvement in beam selection accuracy on real-world NEU dataset in Sec.~\ref{sec:real_world_analysis}.}

% For all models, we exploit categorical cross-entropy loss for training with batch size of 32 for 100 epochs with an earlier stopping point of patience 10. We use Adam~\cite{adam} as optimizer with $\beta=(0.9,0.999)$ and initialize the learning rate to 0.0001. Moreover, we choose rectified linear units (ReLU) as activation function for convolution layers. The activation functions introduce a non-linearity on feature maps that allows models to learn a wide variety of functions. Compared to other activation functions, ReLU sets negative values to zero, which results in sparser outputs. The sparsity may be pursued for multiple reasons in a CNN but it mainly provides robustness to small changes in input. 

\subsection{Performance of Base Unimodal Architectures}
We assess the performance of beam selection by only relying on unimodal data. %As mentioned before, we use the S008 dataset for our train and validation set and S009 as the test set. 
First, we preprocess image and LiDAR raw data using the methods proposed in Sec.~\ref{sec:preprocessing}. Then, we normalize and feed it to corresponding base unimodal architectures presented in Fig.~\ref{fig:model_architectures} followed by a softmax activation at the output layer. The experimental results of predicting top-$K$ beam pairs are presented in Tab.~\ref{tab:results}, for each proposed unimodal architectures. In the table, we report the top-$K$ ($K$=1, 2, 5, 10, 25, 50) accuracy along with weighted recall, precision and F1 score and the KL divergence of the predicted labels and true labels {\color{black} on Raymobtime dataset}. We observe that the LiDAR outperforms coordinate and image in all metrics with 46.23\% top-1 accuracy, which makes it the best single modality. {\color{black}Moreover, to justify the improvement achieved by using the image preprocessing step described in Appendix~\ref{appendix:imgae_preprocessing}, we compare the weighted recall on raw and preprocessed image data. Interestingly, we observed that by using the raw images, the model always predicts the class with the highest occurrence in the training set that results in the weighted recall of 0.01\%. Intuitively, in the case of using raw images, the model cannot find a relation between the input image and the labels since from a raw image perspective any vehicle captured in the image can be the target receiver. On the other hand, using the image preprocessing step increases the weighted recall to 7\% as presented in Tab.~\ref{tab:results}.}
\vspace{-2mm}
\subsection{Performance of Fusion Framework}
\begin{figure}
    \centering
    \includegraphics[width=0.75\linewidth]{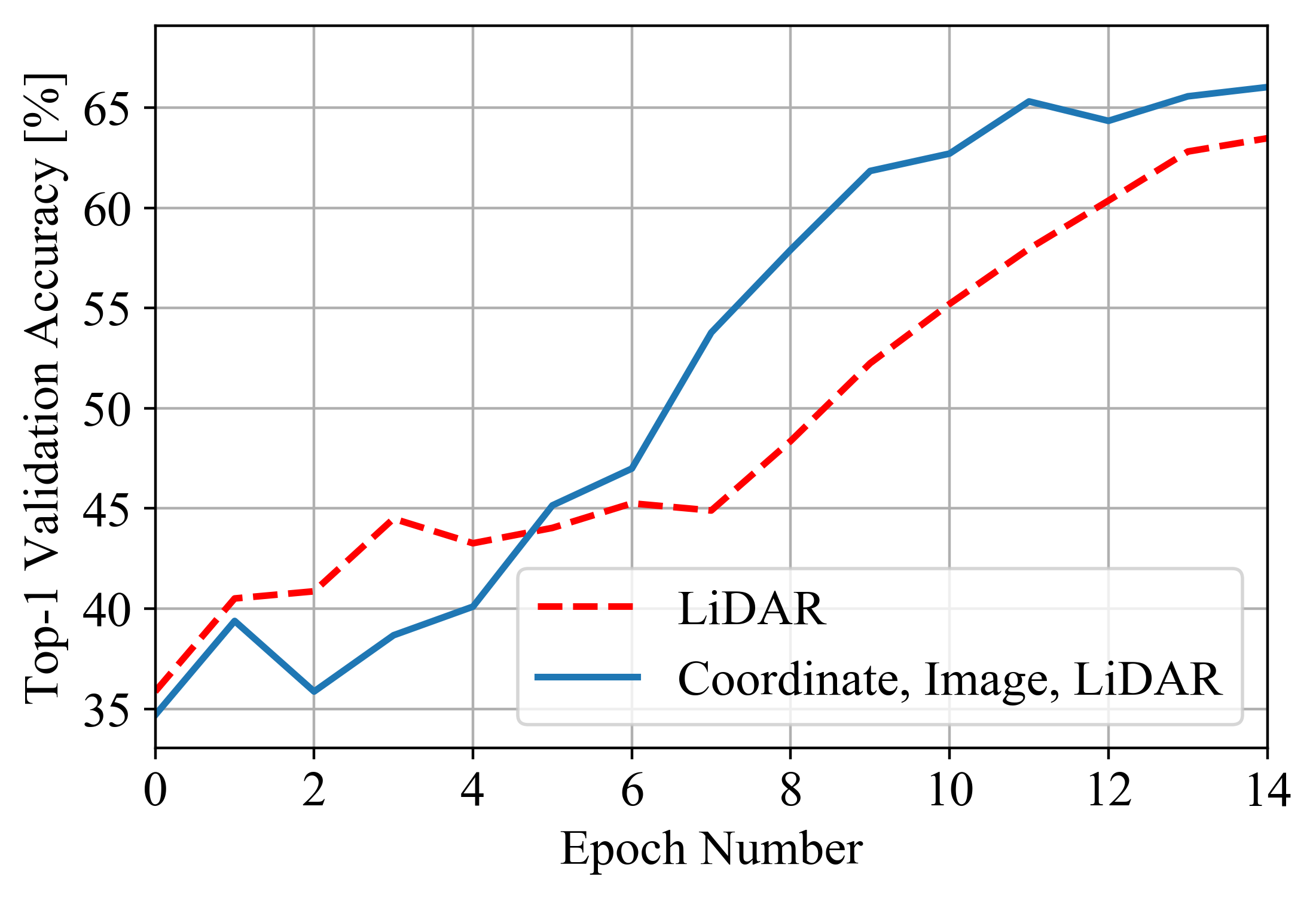}
    \caption{Comparing top-1 validation accuracies of LiDAR-only and fusion with all three modalities {\color{black} on the Raymobtime dataset}.}
    \label{fig:fusion-all-train-val}
    \vspace*{-10pt}
\end{figure}

The results of fusion on different combinations of unimodal data are presented in Tab. \ref{tab:results} {\color{black} for Raymobtime dataset}. We observe that the fusion increases the beam prediction accuracy in all combinations. Moreover, the best result is achieved when all modalities are fused together with $9.99\%$ improvement in top-1 accuracy in comparison with the best unimodal data i.e., LiDAR. The improvement with fusion can be also justified by tracking the validation accuracy during training. Fig. \ref{fig:fusion-all-train-val} compares the top-$1$ validation accuracy of fusion of all three modalities with LiDAR-only (best single modality). %From this figure, 
We observe that although the top-$1$ validation accuracy of fusion is lower in early epochs, it outperforms the LiDAR after five epochs.

Since the dataset is highly imbalanced, we report results using metrics like weighted precision, recall, and F1 score to confirm the improvement. Furthermore, we use KL divergence metric to measure the overall performance of the fusion pipeline. The lower the divergence, the more is the similarity between true and predicted labels. We also use KL divergence to show the relative entropy between train (S008) and test (S009) data labels (Shown in Fig. \ref{fig:dist_s89}). We get KL divergence of 0.57 signifying high relative entropy between the train/test label distributions. From Tab.~\ref{tab:results}, we observe that the fusion with all unimodal data leads to the lowest KL scores. Hence, we deduce that fusion among all three modalities is the most successful scheme to capture the label distribution in the test set. 
% It can also be observed by comparing the distribution of the test labels (Fig. \ref{fig:s9_frequency}) and predicted labels by the model (Fig. \ref{fig:hm_test_pred}).  \\removed
Hence, we choose the proposed fusion-based approach comprising of all three modalities as beam selector for the rest of the performance evaluation. %KRC- did we claim that we intelligently decide which modality must be combined? Or is it a blanket statement that says, "use all sensors"

\vspace{-2mm}
\subsection{Studying the Impact of $K$}
\begin{figure*}[ht]
\begin{subfigure}{0.3\textwidth}
  \centering
  \includegraphics[width=\linewidth]{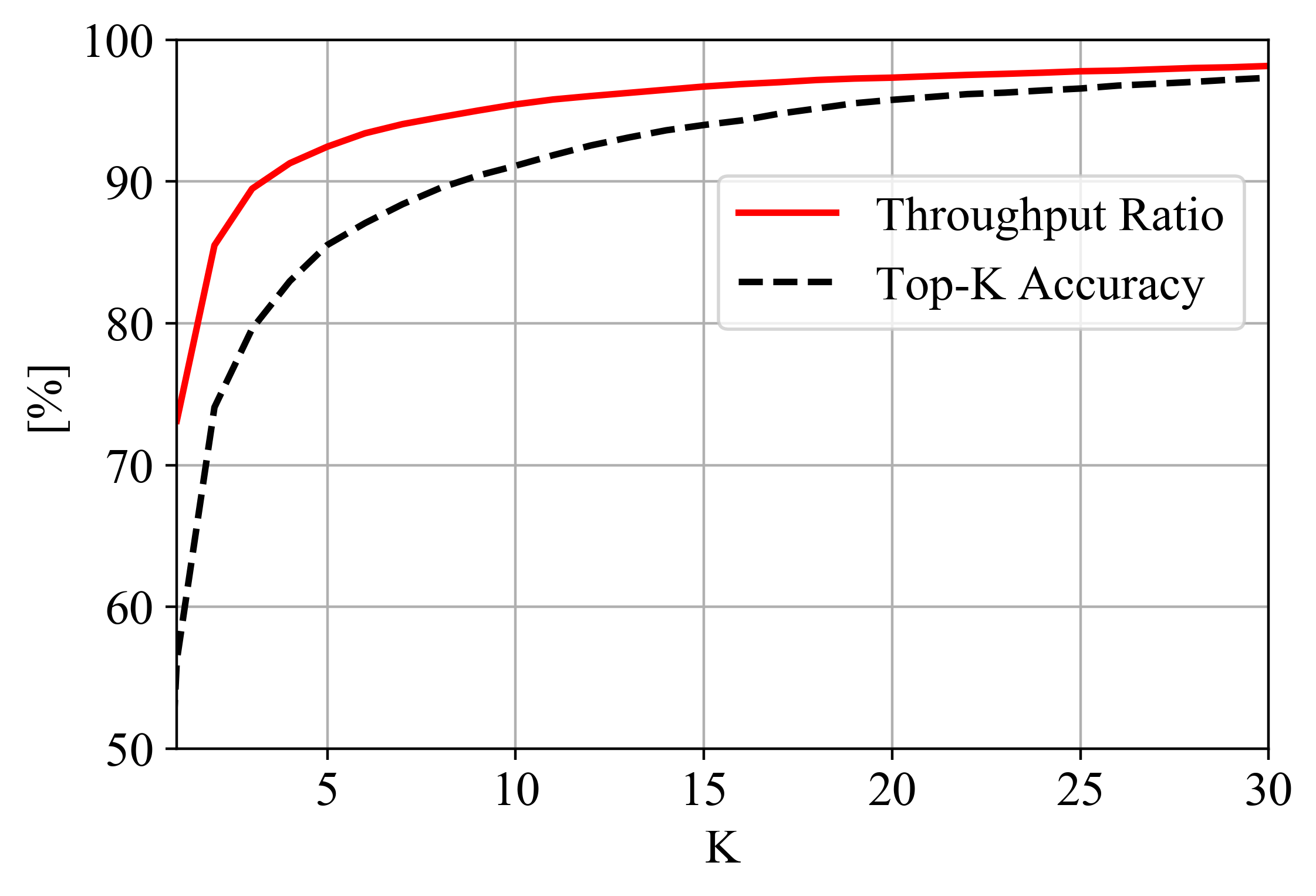}  
  \caption{}
  \label{fig:throuput_ratio}
\end{subfigure}
\hspace{2mm}
\begin{subfigure}{0.3\textwidth}
  \centering
  \includegraphics[width=\linewidth]{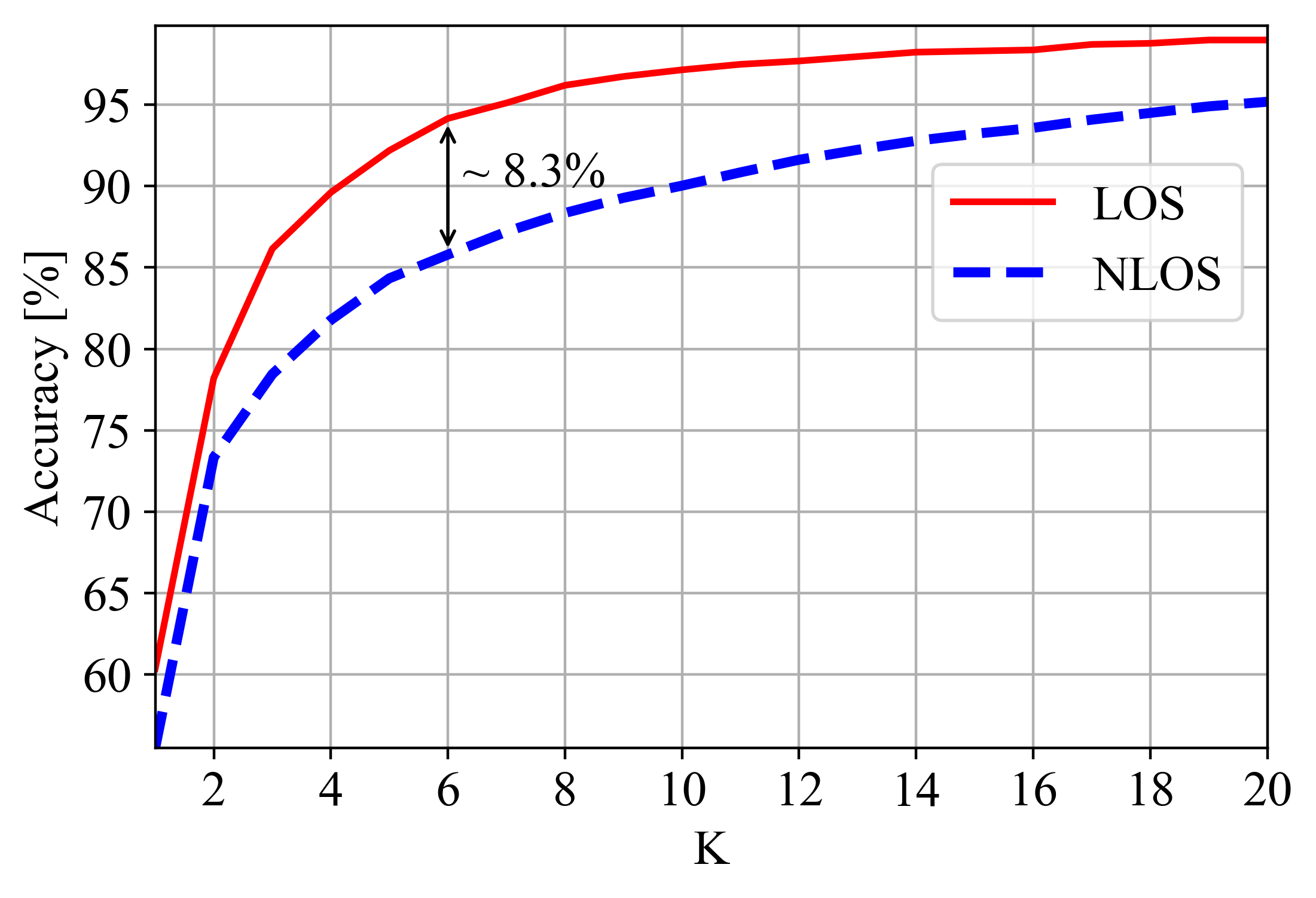}  
  \caption{}
  \label{fig:los/nlos_accuracy}
\end{subfigure}
\hspace{2mm}
\begin{subfigure}{0.36\textwidth}
  \centering
  \includegraphics[scale=0.42]{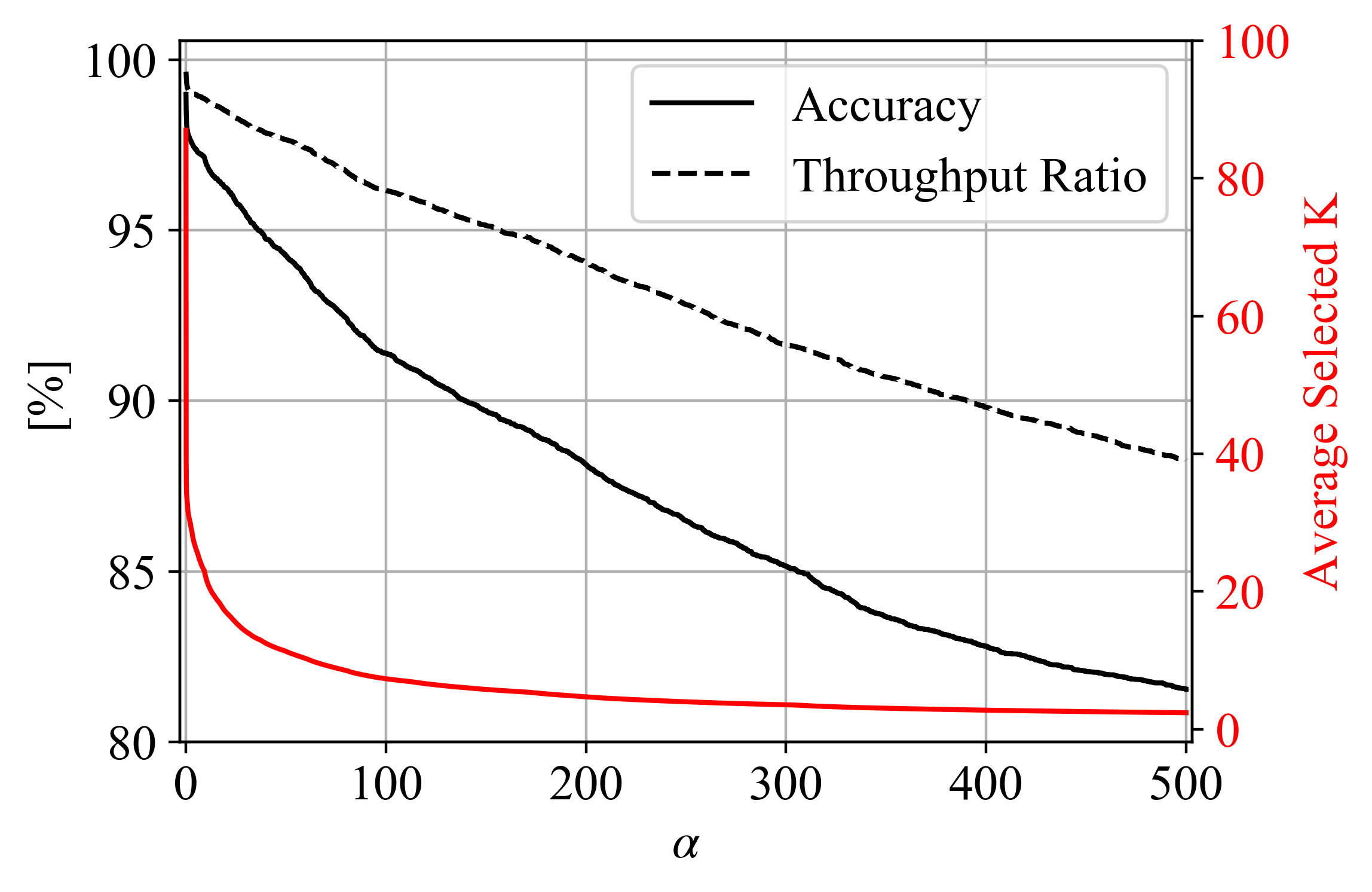}  
  \caption{}
  \label{fig:K_selection}
\end{subfigure}
\caption{(a) Comparison of throughput ratio and beam selection accuracy with varying $K$ (b) LOS/NLOS accuracy for $K=0,1,...,20$ (c) Analysis of throughput ratio, accuracy and average selected $K$ for different $\alpha$ values in Eq. \eqref{eq:P_KgeqK}.}
\label{fig:result_three}
\vspace*{-10pt}
\end{figure*}

{\color{black}To analyze the impact of different $K$ values in the overall performance, we point out that failure in selecting the optimum beam pair within the suggested subset ($(t^*,r^*) \notin \mathcal{B}_k$) results in the drop in the received signal power. Hence, we choose the throughput ratio~(see Appendix B) as our metric to assess the QoS of the system. Intuitively, the throughput ratio depicts the ratio of average throughput when sweeping only $K$ beam pairs predicted by the model with reference to what could be achieved with exhaustive search.} Fig.~\ref{fig:throuput_ratio} compares the throughput ratio and normalized beam selection accuracy with $K$ varying from 1 to 30 {\color{black} for Raymobtime dataset}. As expected, both increase with $K$ since it is more likely to include the optimum beam pair with higher $K$. We observe the gap between the accuracy and throughput ratio starts with 16.90\% for $K$=1, and it decreases as $K$ increases. We do not observe significant improvement in throughput ratio after $K = 10$; however, the accuracy keeps on improving until $K = 25$. Note that while increasing $K$ improves the quality of service~(QoS), it results in higher beam selection overhead as well. Hence, it is crucial to balance the tradeoff between the two as proposed in dynamic selection of top-$K$ beam pairs algorithm in Sec.~\ref{sec:dynamic_k_selection}.

% and a beam selection task with 256 beam pairs.  \\removed \\was in caption

% \begin{figure}[t]
%     \centering
%     \includegraphics[width=0.8\linewidth]{figs/Results/accuracy_and_throughput_all_samples.png}
%     \caption{Comparison of throughput ratio and beam selection accuracy with varying $K$.}
%     \label{fig:throuput_ratio}
% \end{figure}

\vspace{-2mm}
\subsection{Impact of LOS and NLOS}
The presence of obstacles leads to massive drops in channel quality given the high attenuation in the mmWave band. Additionally, users might experience a considerable reduction in their QoS to tens of Gbps. In the case of LOS scenario, the corresponding best beam pair distinctively outperforms the others. However, the presence of blockage in LOS path causes unexpected beams to achieve the highest signal strength through multiple reflections. We show this in Fig.~\ref{fig:los/nlos_accuracy}, which compares the accuracy of our proposed fusion where the sample of test data are separated based LOS/NLOS scenario {\color{black} in Raymobtime dataset}. As expected, prediction in the case of complex reflections of NLOS links is more challenging, showing a maximum drop of 8.3\% in beam selection accuracy against LOS scenarios.
%in comparison with LoS scenarios.

% \begin{figure}[t]
%     \centering
%     \includegraphics[width=0.8\linewidth]{figs/Results/accuracy_LOS_NLOS.png}
%     \caption{LoS/NLoS accuracy for $K=0,1,...,20$ and a beam selection task with 256 beam pairs.}
%     \label{fig:los/nlos_accuracy}
% \end{figure}

\subsection{Impact on Beam Selection Speed}
\label{subsec:beam_selection_speed}
As discussed in Sec.~\ref{subsec:beamTime}, the 5G-NR standard define a brute-force beam sweeping process that sequentially explores all possible directions. In addition, according to Eq.~\eqref{eq:T_nr}, only up to 32 directions can be explored within one SS burst, which creates additional waiting time within one beam selection process. In order to decrease such overhead, we propose a solution that selects a reduced set of $K$ beam pairs and performs a brute-force search only on those ones. Also, given the different confidence levels of our prediction model due to potential scenario variations, we propose an algorithm that selects $K$ flexibly to avoid unnecessary overhead. 
% https://electronics.stackexchange.com/questions/288530/uniform-linear-array-ula-beamwidth-and-angular-resolution-using-fft

In the Raymobtime dataset, the road length is 200 meters and the BS is located in the middle. On the other hand, the 3-dB beam width of an uniform linear array antenna with $N$ elements is approximately equal to $2/N$ radians~\cite{richards2014fundamentals} that results in span of ${3.58}^{\circ}$ and ${14.32}^{\circ}$ for each beam of transmitter and receiver codebooks, respectively. Hence, the overall BS coverage angle is equal to $\phi_{BS} = {114.56}^{\circ}$ and the contact time, i.e., the time that the vehicle remains in the span of one beam, is equal to $T_{total} = \frac{2h~\tan(\frac{\phi_{BS}}{2})}{v_l}$ with $h$ and $v_l$ being the height of the BS and the velocity of the vehicle \cite{reusmuns2019beam}. Consequently, the vehicle remains in the coverage region of each beam pair for $\sim 807\,ms$ while moving with the velocity of 32 km/h (average speed in urban roads). Therefore, the beam selection process needs to be repeated every $807\,ms$ ($T_{total}$). In Fig.~\ref{fig:K_selection}, we analyze the impact of $\alpha$ in Eq. \eqref{eq:P_KgeqK} on the throughput ratio~($R_T$), the accuracy and the average selected $K$. We observe how the triplet {\em $R_T$, accuracy} and {\em average selected $K$} decreases with $\alpha$, the control parameter in Eq.~\eqref{eq:P_KgeqK}. Intuitively, increasing $\alpha$ gives more weight to the second term in Eq. \eqref{eq:P_KgeqK} that forces the algorithm to be faster and choose lower $K$ which results in lower QoS and beam selection accuracy. Interestingly, we observe that for $\alpha=0$ the maximum average selected $K$ is equal to $87$. In this scenario, the objective in Eq. \eqref{eq:P_KgeqK} aims to maximize the alignment probability and increasing the $K$ and yet it does not exceeds $87$ {\color{black} out of 256}. We conclude that our proposed fusion method achieves to $\sim$100\% top-$87$ accuracy, so it does not need to sweep any further beam pairs.

\begin{figure}[t!]
    \centering
    \includegraphics[width=0.9\linewidth]{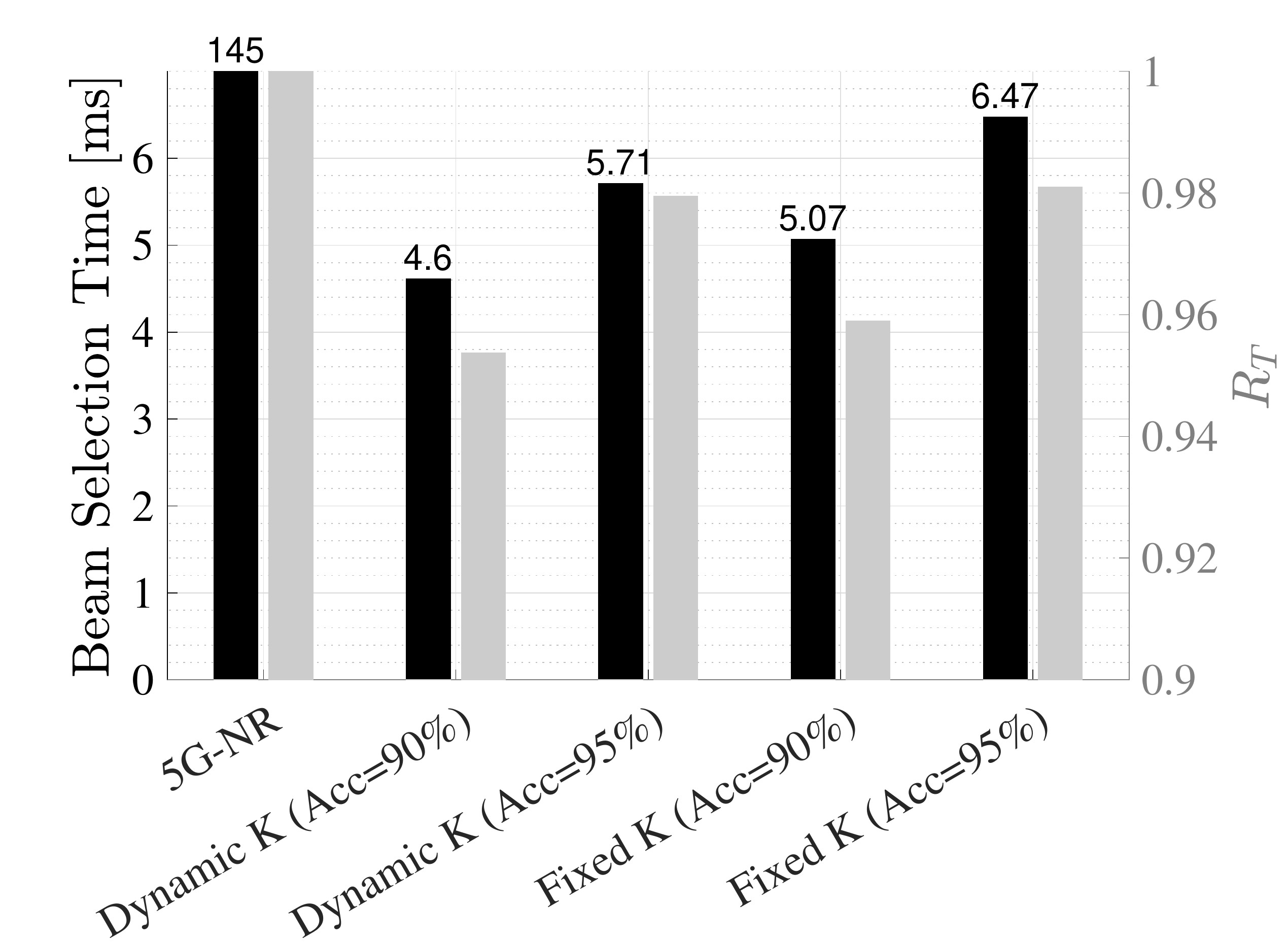}
    \caption{Comparison of relative throughput and {\color{black}end-to-end} beam selection time~{\color{black}({\it Eq.~(\ref{eq:T_df}}))} of proposed approaches, Dynamic~$K$~{\color{black}({\it Algorithm~\ref{alg:K-selection}})} and Fixed~$K$~({\it Eq.~(\ref{eq:best_beam_pair}})), with 5G-NR standard. The actual beam selection time of $145\,ms$ for 5G-NR is scaled here, for better visibility and comparison purpose.}
    \label{fig:TimeComparison}
    \vspace*{-5pt}
\end{figure}

The control parameter in Eq. \eqref{eq:P_KgeqK} enables us to slide between different accuracy and overhead conditions. Fig.~\ref{fig:TimeComparison} shows that the dynamic $K$ selection approach achieves an average throughput ratio of 95.37\% and 97.95\% while targeting 90\% and 95\% accuracy, respectively.{ \color{black} This implies that the capacity of the proposed F-DL approach is only 4.63\% lower than the 5G-NR standard, while targeting the accuracy of 90\% for instance}. Moreover, the dynamic $K$ selection approach offers the corresponding beam sweeping overhead of $0.94\,ms$ and $2.04\,ms$, Eq.~\eqref{eq:T_sweep} and the overall beam selection delay of $4.6\,ms$ and $5.71\,ms$. {\color{black} Note that the beam selection delay of our proposed dynamic $K$ selection method in Fig.~\ref{fig:TimeComparison} corresponds to the end-to-end latency of the proposed F-DL method presented in Eq.~\eqref{eq:T_df}}. In contrast, the 5G-NR standard beam selection procedure requires $145\,ms$. Therefore, we notice 96\% reduction in overall beam selection overhead while retaining 97.95\% relative throughput associated with 95\% accuracy. 
% Furthermore, we compare the performance of proposed dynamic $K$ selection with the fixed $K$ one (Fig. \ref{fig:throuput_ratio}).
Furthermore, we compare the performance of proposed algorithm for constructing the subset $\mathcal{B}_K$, Algorithm~\ref{alg:K-selection}, that is generated \textit{dynamically} per case, with the fixed $K$ one (Fig.~\ref{fig:throuput_ratio}). Note, that fixed $K$ selection is a posterior probability derived after observing all test samples; however, the dynamic $K$ selection selects the $K$ for each sample of test set, independently. From this figure, we observe that the proposed dynamic $K$ selection approach outperforms the fixed $K$ one, providing faster beam selection with close competing relative throughput while targeting the same accuracy. We use the same standard, i.e., 5G-NR for fair comparison (see  Fig.~\ref{fig:TimeComparison}). Note that our algorithm can be trivially extended to any other exhaustive beam search standards, such as IEEE 802.11ad by modifying Eq.~\eqref{eq:T_sweep}, yet it does not negate the improvement achieved by restricting the beam selection to a lower dimension space. 
% \vspace{-2mm}

{\color{black}
\subsection{Real-world Implementation}
\label{sec:real_world_analysis}
We validate the performance of the proposed fusion deep learning method on the home-grown NEU dataset. As mentioned in Sec.~\ref{sec:dataset_real}, due to the infrastructural limitation, we use only LiDAR and GPS branch of the proposed F-DL~(presented in Sec.~\ref{sec:proposed_fusion}, Fig.~\ref{fig:fusion_pipeline}) for this set of experiment.
% We discard the image branch of the proposed fusion deep learning method presented in Sec.~\ref{sec:proposed_fusion} and use GPS and LiDAR sensor inputs to configure the best beam pair.
Tab.~\ref{tab:FLASH_fusion_results} compares the beam selection accuracy while using individual sensor inputs in contrast to the case where the information from GPS and LiDAR sensor are fused together. We observe that fusion improves the Top-1 prediction accuracy from 74.86\% for the best modality, i.e., LiDAR to 78.18\% for the fusion of GPS and LiDAR sensors. The weighted F1 score also increases by 3.6\% denoting better handling of imbalances in ground-truth, which is common in mmWave beams. 
}

{\color{black}
\subsection{Accuracy and End-to-End Latency Analysis:}
The Raymobtime and NEU datasets have 256 and 64 possible beam pairs each; hence, sweeping the entire codebook elements requires, $145\,ms$ and $25\,ms$, respectively, according to 5G-NR standard explained in Sec.~\ref{subsec:beamTime}. On the other hand, the proposed beam selection method restricts the beam search space to a subset of $K$ beam pairs. We study the trade-off between the accuracy and end-to-end beam selection time~(presented in Eq.~\eqref{eq:T_df}) versus $K$ in Fig.~\ref{fig:acc_time_K} for both datasets. In particular, we observe that for the Raymobtime dataset the accuracy is $>99\%$ for $K> 87$ while the end-to-end latency is still increasing. On the other hand, for the NEU dataset, the accuracy and end-to-end beam selection time starts with 78.18\% and $3.818\,ms$ for $K=1$. The accuracy saturates at $K=7$ and reaches $\sim100\%$ for $K>12$ while the beam selection time keeps on increasing and becomes $25.86\,ms$ for $K=64$. Specifically, Fig.~\ref{fig:acc_time_K} highlights the importance of the $K$ selection method to choose the appropriate $K$ and avoid unnecessary overhead imposed on the system. 
\begin{figure}[ht]
\begin{subfigure}{0.23\textwidth}
  \centering
  \includegraphics[width=\linewidth]{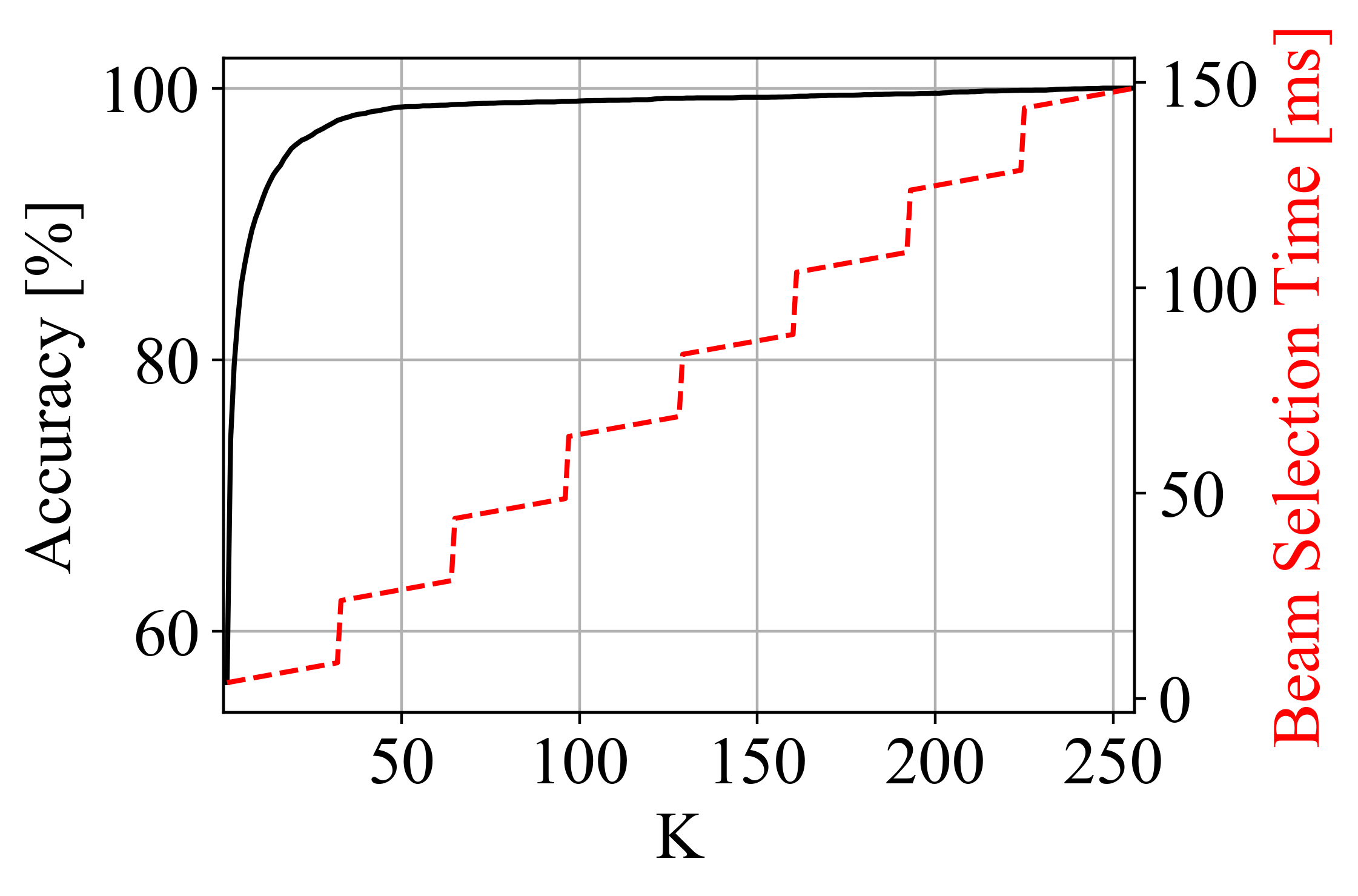}  
  \caption{}
  \label{fig:raymobtime_acc_time}
\end{subfigure}
\hspace{2mm}
\begin{subfigure}{0.23\textwidth}
  \centering
  \includegraphics[width=\linewidth]{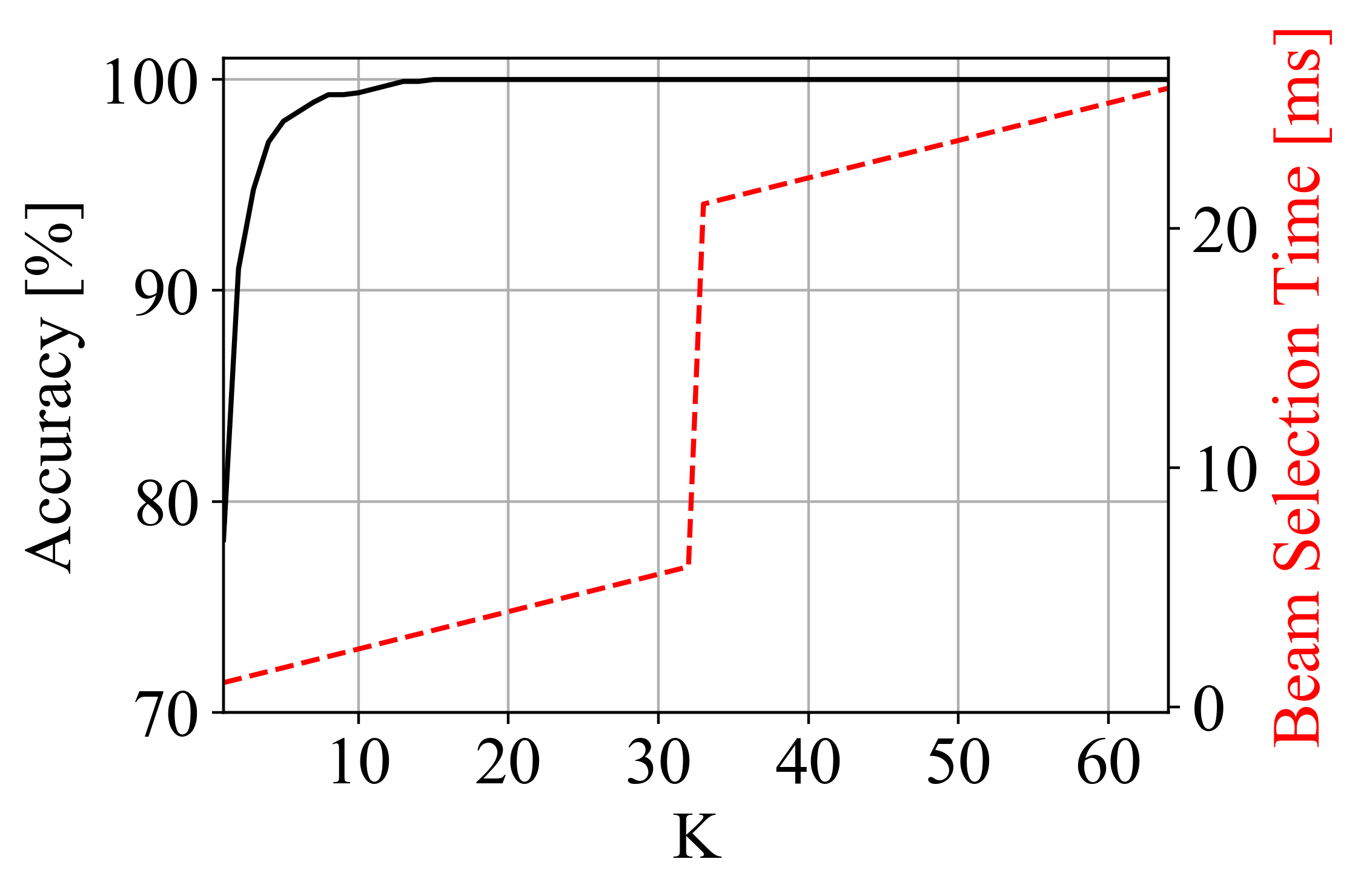}  
  \caption{}
  \label{fig:Raymobtime_FLASH_acc_time}
\end{subfigure}
\hspace{2mm}
\caption{{\color{black}Beam selection accuracy and end-to-end beam selection time versus K on the (a) Raymobtime and (b) NEU datasets}.}
\label{fig:acc_time_K}
\vspace*{-5pt}
\end{figure}

}
\begin{table}[t!]
    \setlength{\tabcolsep}{5pt}
    \centering
\resizebox{0.45\textwidth}{!}{{\color{black}
     \begin{tabular}{|c|c|c|c|c||} 
    %  \arrayrulecolor{blue}
     \hline 
    Modalities & Top-$1$ & Top-$2$ & Top-$5$ & Weighted \\
      & Accuracy & Accuracy & Accuracy & F1 score  \\
     \hline \hline
    Coordinates & 39.94\% & 54.39\%  & 81.05\% & 33.63\% \\ 
     \hline
    LiDAR & 74.86\% & 89.04\%& 97.57\%  & 75.02\%  \\
     \hline
    Coordinates, LiDAR& 78.18\% &  91.02\% & 98.02\% & 78.62\% \\
    \hline
    \end{tabular}}}
        \caption{{\color{black}Performance of proposed unimodal and fusion method on real-world NEU dataset.}}
    \label{tab:FLASH_fusion_results}
    \vspace*{-5pt}
\end{table}

% \begin{figure}
%     \centering
%     \includegraphics[width=0.75\linewidth]{figs/Results/FLASH_acc_time.png}
%     \caption{{\color{blue}Top-1 accuracy of GPS and LiDAR fusion and end-to-end beam selection time versus K on NEU dataset} {\color{red}add the raymbtime one}}
%     \label{fig:FLASH_acc_time}
% \end{figure}

{\color{black}
\subsection{Comparison with the State-of-the-art}

In Tab.~\ref{tab:comparison}, we compare the performance of our proposed models to the state-of-the-art DL based approaches by Klautau {\em et al.}~\cite{klautau2019} and Dias {\em et al.}~\cite{dias2019position}, both evaluated on the Raymobtime dataset. To the best of our knowledge, these are the only methods that include equivalent scenarios to the ones considered in this paper. In particular, LiDAR sensor data collected on vehicles is used for beam prediction under both LOS and NLOS conditions. Other works that consider different evaluation metrics (\cite{va2017inverse, wang2018, va2016beam, Aviles_2016}), camera images under LOS-only scenarios (\cite{alrabeiah2020viwi,tian2020applying,xu20203d}) or RF data \cite{alrabeiah2020millimeter} have been kept out of the comparison. As we show in Tab.~\ref{tab:comparison}, the proposed LiDAR model and the F-DL architecture outperform the state-of-the-art (\cite{klautau2019, dias2019position}) by 18.95-20.45\% and 20.11-21.61\% respectively in top-$10$ accuracy. {\color{black}Moreover, both Klautau {\em et al.}~\cite{klautau2019} and Dias {\em et~al.}~\cite{dias2019position} only keep the codebooks elements that are at least seen 100 times in the training set.}

% In Tab.~\ref{tab:comparison}, we benchmark the performance of our proposed models against the state-of-the-art DL based approaches by Klautau {\em et al}~\cite{klautau2019} and Dias {\em et al.}~\cite{dias2019position}, where the authors used Raymobtime datasets as well. Both these techniques used centralized inference with LiDAR sensor at the vehicle considering both LoS and NLoS situations. However, we limit the comparison study to only these aforementioned techniques as the other state-of-the-art differ from our evaluation with respect to various aspects, such as: (a)~different evaluation metrics~\cite{va2017inverse, wang2018, va2016beam, Aviles_2016}, (b)~consideration of LOS-only scenarios while using camera sensors~\cite{alrabeiah2020viwi,tian2020applying,xu20203d}, (c)~inclusion the RF inputs (sub-6~GHz channel measurements for instance)~\cite{alrabeiah2020millimeter}. 
% In Tab.~\ref{tab:comparison}, we observe that our proposed LiDAR model and F-DL architecture outperform the state-of-the-art by 18.95-20.45\% and 20.11-21.61\% respectively in top-$10$ accuracy. 
%F-DL also provides the extended support to incorporate the GPS and image as modalities in our designed fusion framework that results in improved performance.
\begin{table*}[hbtp]
    \setlength{\tabcolsep}{5pt}
    \centering
\resizebox{0.85\textwidth}{!}{
     \begin{tabular}{|c|c|c|c|c|c|c|c|c|} 
     \hline 
    Methods & \textcolor{black}{Dataset} & \textcolor{black}{\# Beams} & Modalities  & Inference & Top-$1$ & Top-$2$ & Top-$5$ &Top-$10$\\ 
    \hline \hline
    Dias {\em et al.} \cite{dias2019position} & \textcolor{black}{Raymobtime (S007)} & \textcolor{black}{264} & LiDAR & Centralized & $20.5\pm 1 \%$ & $25.5\pm 1 \%$ & $54.5\pm 1 \%$ & $68.5\pm 1 \%$ \\
    \hline
    Klautau {\em et al.} \cite{klautau2019} & \textcolor{black}{Raymobtime (S008)} & \textcolor{black}{240} & LiDAR & Centralized & $30.5\pm 1$\% & $43.5\pm 1 \%$ & $57.5\pm 1 \%$ & $70\pm 1 \%$\\
    \hline
    \hline
    Proposed LiDAR Network & \textcolor{black}{Raymobtime~(S008)} & \textcolor{black}{256} & LiDAR & Centralized & $46.23\%$ & $64.67 \%$ & $82.43 \%$ & $89.95 \%$ \\
    \hline
    Proposed F-DL & \textcolor{black}{Raymobtime~(S008)} & \textcolor{black}{256} & GPS, Image, LiDAR & Distributed & {\bf 56.22\%} & {\bf 74.08\%} & {\bf 85.53\%} & {\bf 91.11\%} \\
    \cline{2-9}
    & \textcolor{black}{NEU} & \textcolor{black}{64} & \textcolor{black}{GPS, LiDAR} & \textcolor{black}{Distributed} & \textcolor{black}{{\bf 78.18\%}} & \textcolor{black}{{\bf 91.02\%}} & \textcolor{black}{{\bf 98.02\%}} & \textcolor{black}{{\bf 99.37\%}} \\
    \hline
    \end{tabular}}
        \caption{Comparison of proposed best performing unimodal and F-DL architectures with two benchmark DL based approaches {\color{black} on Raymobtime dataset~\cite{klautau20185g} and results on the real-world NEU dataset.}}
    \label{tab:comparison}
    \vspace*{-10pt}
\end{table*}
}

\subsection{Discussion}
%discuss about then to use what. In case of missing sensor data, the network will
We summarize below interesting observations from the experimental results:
\begin{itemize}
    \item %Even though incorporating image data with LiDAR and coordinate fusion does not give significant performance boost, it helps us in two folds. First of all wireless channel is fraught with uncertainties, and the instability even increases for higher frequency bands. Adverse wireless environment often results in missing information. 
    When LiDAR and GPS sensors are deployed over the vehicle and data is transmitted to the BS through sub-6~GHz data channel, the wireless control channel may impact the actual delivery at the MEC. On the other hand, cameras at the BS may have a reliable fiber connectivity to the MEC. Hence, in case of unreliable channel conditions or faulty sensors, our fusion framework is still able to make predictions 
    %on the beams pair selection  \\remove
    based on any available sensor modality. This robustness to unreliable channel conditions is essential, even if there is no immediate gain from fusing a specific type of modality.
    % Also, the image sensor, being deployed in BS, does not add latency to the overall framework. 
    %Hence, we argue that the proposed fusion-based beam selector is resilient to unreliable network condition.
    \item Proposed beam selection technique with dynamically chosen $K$ automatically selects the top-$K$ best beam pairs, with performance closed to a fixed $K$ when the latter is identified via expert knowledge. Thus our approach eliminates the need to include expert domain knowledge (know what $K$ is needed to achieve certain amount of accuracy), by automating the %complete 
    beam selection process.
    \item We show that it is possible to reduce the beam-selection overhead in a practical and emerging 5G-NR standard by 95--96\%, while maintaining 97.95\% relative throughput.
\end{itemize}

\vspace{-2mm}
\section{Conclusions}
Increasing softwarization and ability to automatically configure parameters~\cite{Kai_2021} within generation 5G and beyond networks will necessitate the use of ML-based methods distributed at the MEC. In this paper, we propose an approach for ML-aided fast beam selection technique, where multimodal non-RF sensor data is exploited to reduce the search space for identifying best performing mmWave beam beams. Our proposed fusion method exploits the latent embeddings from each unimodal feature representation and the overall framework is evaluated in realistic emulated settings. {\color{black} We observe around 20-22\% increase in performance for top-10 accuracy than the state-of-the-art using the proposed F-DL architecture.} We {\color{black} also} achieve 95--96\% decrease in beam selection time compared to the exhaustive search defined by the 5G-NR standard in the high-mobility urban scenarios. We propose to extend this framework ahead to multiple-receiver scenarios, incorporate federated learning among the sensors, implement network compression and pruning for feasible deployment over IoT edge devices. 
\appendices\
\vspace{-2mm}
\section{Object Detection Algorithm}
\label{appendix:imgae_preprocessing}

{\color{black}Our proposed image preprocessing step is a combination of a standard multi-object detection approach followed by a refinement step where each detected object is denoted by a unique indicator according to their role, i.e., target receiver or obstacle.} 
% We employ a {\color{black} mulit-object detection approach that enables us to flexibly handle the input images}. 
It is constituted of a classifier that is capable to predict the presence of objects in the small bounding boxes. In the training phase, we separately label the examples from the valid items in the environment. We then quantize the samples by filtering the images with a moving square-shaped window of size $W\times W$ pixels. Starting from the top left side of the image, and after generating the first crop, we move the window by $X$ pixels. This process results in a dataset of cropped samples from each of possible items in the environment. Since the dimensions of items vary, we end up with  different number of samples for each class. To achieve a balanced dataset, we augment the minority classes by applying different light conditions, until we reach the same number of samples per class. 
We split the final balanced dataset in (70\%,15\%,15\%) proportion, and train the classifier. 

Similarly, in testing phase, we quantize the image by sweeping it with a window of dimension $W\times W$ and step size $X$. 
% each image is reduced to $C$ crops as:\vspace{-0.08in}
% \begin{equation}
%     C =\lfloor\frac{H-W}{S}+1\rfloor \times \lfloor\frac{L-W}{S}+1\rfloor,
%     \label{eq:crops}
% \end{equation}
Next, we feed each crop to the trained classifier and arrange the predictions in the same order as the crop generation. 
This process leads to a quantized representation of the image, where each element gives the prediction of the classifier for the object in the corresponding $W\times W$ window. We refer to this representation as the \textit{bit map} of the raw input camera images. Given an input image with dimension $H\times L$, the shape of generated bit map will be $\lfloor\frac{H-W}{X}+1\rfloor \times \lfloor\frac{L-W}{X}+1\rfloor$.

We can refine our bit map further if the specific vehicle type is also transmitted directly by the receiver, as part of the basic safety message in IEEE 802.11p standard for instance. %KRC- if there is a standard that does this, mention it.
Therefore, given the generated bit map and the reported type of the target vehicle, we (i) keep the label of legitimate receiver vehicle type, (ii) map other vehicles to obstacles. This process designates the potential location of the target receiver as well as the location of obstacles with much more information than the raw images. Finally, to address the concern that the image preprocessing may introduce significant delay as it requires multiple forward passes, we convert the trained model to an equivalent fully convolutional network. We have previously explored such an approach in~\cite{salehi2021machine}, which enables us to generate the entire bit map in a single forward pass. 
% \begin{figure}
%     \centering
%     \includegraphics[width=0.85\linewidth]{figs/image_preprocess.pdf}
%     \caption{Image preprocessing pipeline}
%     \label{fig:img_preprocess}
%     % diagram avaiable at: https://drive.google.com/file/d/1obCqri-lv1GFU96aOtXg8cyWCMTDrK30/view?usp=sharing 
% \end{figure}

\vspace{-2mm}
\section{Evaluation Metrics}
\label{app:evalution_metrics}
Top-$K$ accuracy 
%is an appropriate metric for the classification problem with a high number of classes. It \\remove
calculates the percentage of times that the model includes the correct prediction among the top-$K$ probabilities. Formally, given $\phi$ a Boolean predicate, let $\id_{\phi}$ to be 1 if $\phi$ is true, and 0 otherwise. Moreover, given ground-truth beam pair $(t^*,r^*)$, the prediction probability score $S\in\mathbb{R}^{|\mathcal{B}|}$, top-$K$ accuracy is defined as:
\begin{equation}
    \label{eq:top-K-def}
    \resizebox{0.48\textwidth}{!}{$Acc@K = \frac{1}{N_t^{'}} \sum_{l=1}^{N_t^{'}} \id_{((t^*,r^*) \in A' |\underset{A'\subset \{1,...,|\mathcal{B}|\},|A'|=K}{\arg\max}~\sum_{j\in A'} s_j )}$}.
\end{equation}
% \begin{equation}
%     \begin{aligned}
%     Acc@K &= \frac{1}{L} \sum_{l=1}^L \id_{(i^* \in \{ \hat{\mathbf{i}} \in \mathbb{R}^{k} | \arg\max_k \hat{\p}_l \})},
%     \end{aligned}
% \end{equation}
where $N_t^{'}$ denotes the number of test samples. Note that for $K=1$ we get the conventional top-1 accuracy that only the highest probability prediction is taken into account.

The F1 score measures a model's ability to perform with imbalanced class distribution. The F1 score is the harmonic mean of precision and recall given as $F1 = 2 \times \frac{\mathrm{precision} \times \mathrm{recall}}{\mathrm{precision} + \mathrm{recall}}$. 
%\begin{equation}
%    \label{eq:F1}
%    \begin{aligned}
%    F1 &= 2 \times \frac{precision \times %recall}{precision + recall},
%    \end{aligned}
%\end{equation}
Precision denotes how many of the predicted true labels are actually in the ground-truth, while recall denotes how many of the actual labels are predicted.
Here, to combine the per-class F1 scores into a multi-class version, we weight the F1-score of each class by the number of samples from that class. Weighted precision and recall are also calculated in similar manner.

KL divergence measures the divergence of the predicted probability distribution from the true one. Given the one-hot encoding $y \in \mathbb{R}^{|\mathcal{B}|}$ of the ground-truth labels and the prediction $\hat{y}$, KL divergence is defined as $ KL(\hat{y} || y) = \sum_{i=1}^{|\mathcal{B}|} \{\hat{y}_i \log \frac{\hat{y}_i}{y_i} \}.$
%\begin{equation}
%    \label{eq:KL_div}
%    \begin{aligned}
%    KL(\hat{y} || y) &= \sum_{i=1}^{|\mathcal{B}|} \{\hat{y}_i \log \frac{\hat{y}_i}{y_i} \}.
%    \end{aligned}
%\end{equation}

% \changed
{\color{black}Finally, we evaluate the performance of our fusion based beam selector with respect to achieved throughput ratio that is defined as $R_T = \frac{1}{N_t^{'}}\sum_{n=1}^{N_t^{'}}\frac{ \log _2[1+y_{\widehat{(t^*,r^*)}}(n)]}{\log _2[1+y_{(t^*,r^*)}(n)]}$,  
%\begin{equation}
%    \label{eq:TR}
%    R_T = \frac{1}{N_t^{'}}\sum_{n=1}^{N_t^{'}}\frac{ log_2[1+y_{\widehat{(t^*,r^*)}}(n)]}{log_2[1+y_{(t^*,r^*)}(n)]},
%\end{equation}
where $(t^*,r^*)$ and $\widehat{(t^*,r^*)}$ show the best beam pair in $\mathcal{B}$ and $\mathcal{B}_k$ (as defined in Sec.~\ref{sec:problem_formulation_802.11ad} and \ref{Sec:sub-set-selection}), respectively, and $N_t$ is the total number of test samples.} 

%\section*{Acknowledgements}
%The authors gratefully acknowledge support by the National Science Foundation (grant CCF-1937500).
\bibliographystyle{IEEEtran}
\bibliography{ref}
%%%%%%%%%%%%%%%%%%%%%%%%%%%bios
\vskip -2.5\baselineskip plus -1fil
\begin{IEEEbiography}
[{\includegraphics[width=1in,height=1.25in,clip,keepaspectratio]{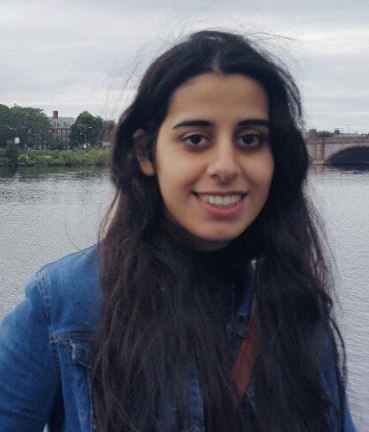}}] 
{Batool Salehi} is currently pursuing a Ph.D. degree in computer engineering at Northeastern University under the supervision of Prof. K. Chowdhury. She received her M.S. (2019) in Electrical Engineering from University of Tehran, Iran. Her current research focuses on mmWave beamforming, Internet of Things, and the application of machine learning in the domain of wireless communication. 
\end{IEEEbiography}

% \vspace{-1cm}
\vskip -2\baselineskip plus -1fil

\begin{IEEEbiography}
[{\includegraphics[width=1in,height=1.25in,clip,keepaspectratio]{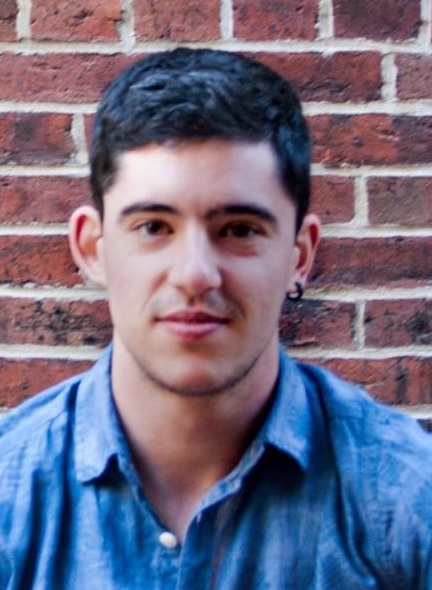}}] 
{Guillem Reus-Muns} received a B.Sc. degree in telecommunications engineering from the Polytechnic University of Catalonia (UPC-BarcelonaTech). He joined Northeastern University, USA, in 2018, where he got his M.Sc. in electrical and computer engineering and is currently working towards his Ph.D. His current research interests involve cellular networks, machine learning for  wireless communications, networked robotics, and spectrum access.
\end{IEEEbiography}
% \vspace{-1cm}
\vskip -2\baselineskip plus -1fil

\begin{IEEEbiography}
[{\includegraphics[width=1in,height=1.5in,clip,keepaspectratio]{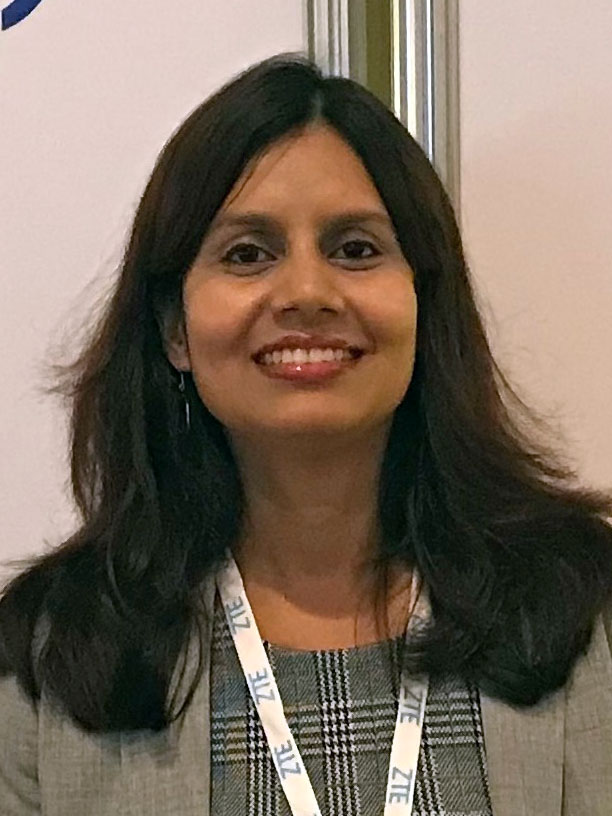}}] 
{Debashri Roy} received her MS (2018) and PhD (2020) degrees in Computer Science from University of Central Florida, USA. She is currently an experiential AI postdoctoral fellow at Northeastern University. Her research interests are in the areas of AI/ML enabled technologies in wireless communication, multimodal data fusion, network orchestration and nextG networks. 
\end{IEEEbiography}
% \vspace{-1cm}
\vskip -2\baselineskip plus -1fil

\begin{IEEEbiography}
[{\includegraphics[width=1in,height=1.25in,clip,keepaspectratio]{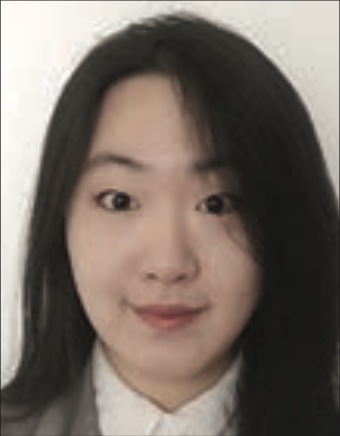}}] 
{Tong Jian} is currently pursuing a Ph.D. degree in the Department of Electrical and Computer Engineering, Northeastern University, Boston, Massachusetts. She received her M.Sc. (2016) in electrical engineering from Rensselaer Polytechnic Institute, New York. She works under the guidance of Prof. Stratis Ioannidis in the field of machine learning. Her current research efforts are focused on the application of machine learning in the domain of wireless communication.
\end{IEEEbiography}
% \vspace{-1cm}
\vskip -2\baselineskip plus -1fil

\begin{IEEEbiography}
[{\includegraphics[width=1in,height=1.25in,clip,keepaspectratio]{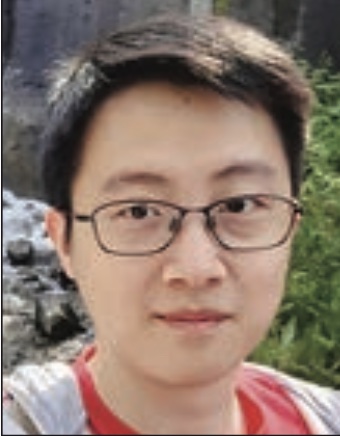}}] 
{Zifeng Wang} is currently pursuing a Ph.D. degree in the Department of Electrical and Computer Engineering, Northeastern University. He received his B.Sc. (2014) in electronic engineering from Tsinghua University, China. He works under the guidance of Prof. Jennifer Dy in machine learning. His current research focuses on lifelong learning, representation learning, and the application of machine learning in the domain of biostatistics and wireless communication. 
\end{IEEEbiography}
% \vspace{-1cm}
\vskip -2\baselineskip plus -1fil

\begin{IEEEbiography}
[{\includegraphics[width=1in,height=1.25in,clip,keepaspectratio]{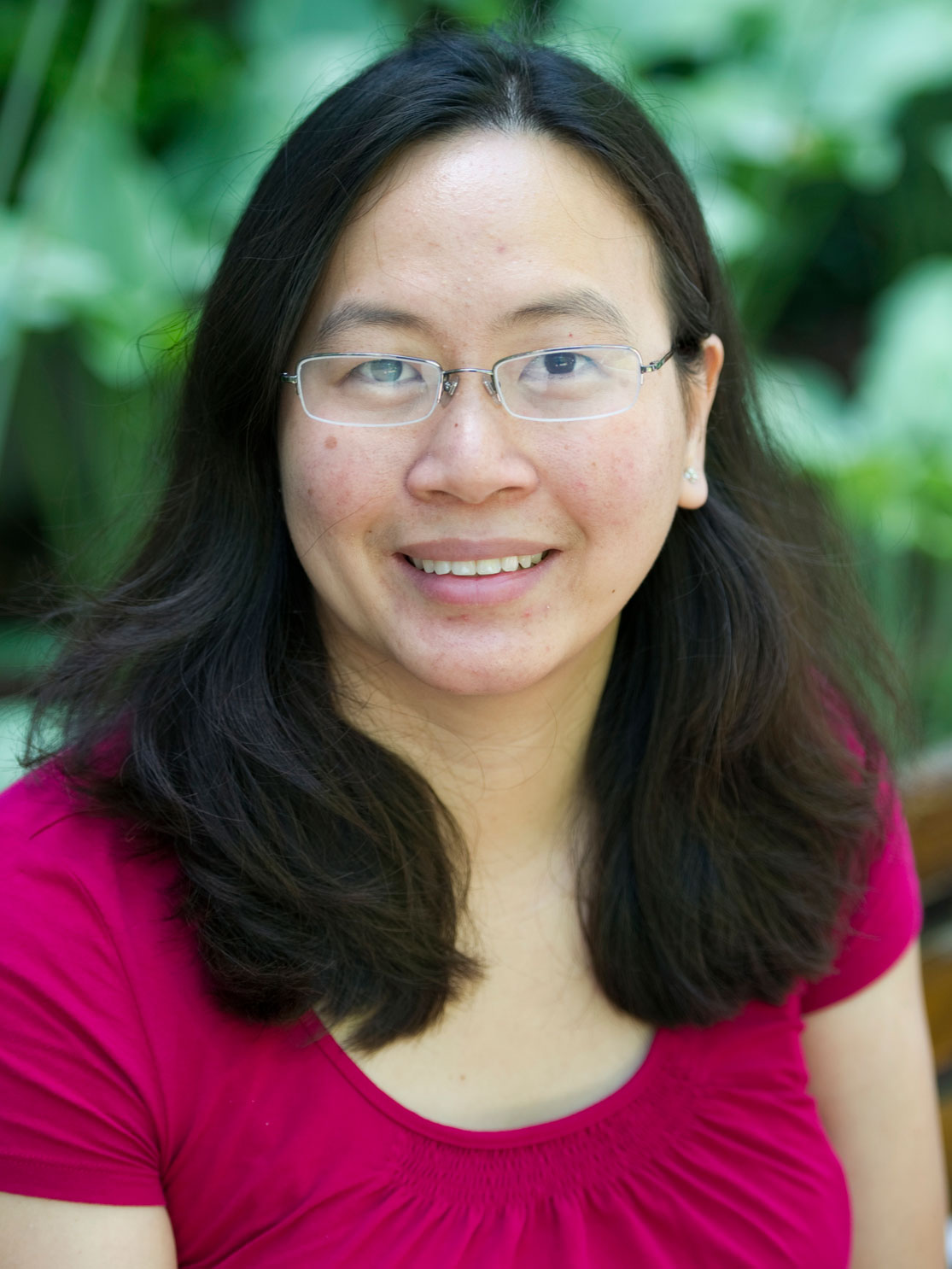}}] 
{Jennifer Dy} is a Professor at the Department of Electrical and Computer Engineering, Northeastern University, Boston, MA, where she first joined the faculty in 2002.  
Her research spans both fundamental research in machine learning and their application to biomedical imaging, health, science and engineering, with research contributions in unsupervised learning, dimensionality reduction, feature selection, learning from uncertain experts, active learning, Bayesian models, and deep representations. 
\end{IEEEbiography}
% \vspace{-1cm}
\vskip -2\baselineskip plus -1fil

\begin{IEEEbiography}
[{\includegraphics[width=1in,height=1.25in,clip,keepaspectratio]{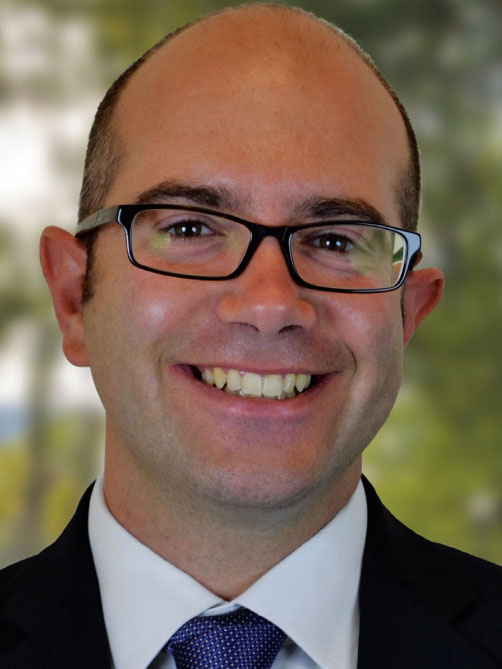}}] 
{Stratis  Ioannidis} is an Associate Professor in the Electrical and Computer Engineering Department of Northeastern University, in Boston, MA, where he also holds a courtesy appointment with the Khoury College of Computer Sciences. 
Prior to joining Northeastern, he was a research scientist at the Technicolor research centers in Paris, France, and Palo Alto, CA, as well as at Yahoo Labs in Sunnyvale, CA. 
His research interests span machine learning, distributed systems, networking, optimization, and privacy.
\end{IEEEbiography}
% \vspace{-1cm}
\vskip -2\baselineskip plus -1fil

\begin{IEEEbiography}
[{\includegraphics[width=1in,height=1.25in,clip,keepaspectratio]{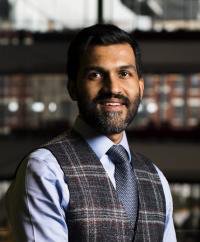}}] {Kaushik Chowdhury} is a Professor at Northeastern University, Boston, MA. 
He is presently a co-director of the Platforms for Advanced Wireless Research (PAWR) project office. His current research interests involve systems aspects of networked robotics, machine learning for agile spectrum sensing/access, wireless energy transfer, and large-scale experimental deployment of emerging wireless technologies.
\end{IEEEbiography}

\end{document}